\pdfoutput=1
\documentclass[10pt,twocolumn,letterpaper]{article}

\usepackage{iccv}
\usepackage{times}
\usepackage{epsfig}
\usepackage{graphicx}
\usepackage{amsmath}
\usepackage{amssymb}

\usepackage[utf8]{inputenc} 
\usepackage[T1]{fontenc}    
\usepackage{url}            
\usepackage{booktabs}       
\usepackage{amsfonts}       
\usepackage{nicefrac}       
\usepackage{microtype}      

\usepackage{caption}
\usepackage{subcaption}
\usepackage{multirow}
\usepackage{xspace}
\usepackage{mathtools}
\usepackage{wasysym}
\usepackage{marvosym}
\usepackage[table,xcdraw]{xcolor}
\usepackage{color}
\usepackage{tabularx}
\usepackage{float}
\usepackage{diagbox,tabu,stackengine}
\usepackage{bbm}
\usepackage{bm}
\usepackage{mwe}
\usepackage{graphbox}

\usepackage{algorithm}
\usepackage[noend]{algpseudocode}

\usepackage{array}
\newcommand{\PreserveBackslash}[1]{\let\temp=\\#1\let\\=\temp}
\newcolumntype{C}[1]{>{\PreserveBackslash\centering}p{#1}}
\newcolumntype{R}[1]{>{\PreserveBackslash\raggedleft}p{#1}}
\newcolumntype{L}[1]{>{\PreserveBackslash\raggedright}p{#1}}

\newcommand{\tabincell}[2]{\begin{tabular}{@{}#1@{}}#2\end{tabular}}

\usepackage[pagebackref=true,breaklinks=true,letterpaper=true,colorlinks,bookmarks=false]{hyperref}

\iccvfinalcopy 

\def\iccvPaperID{5975} 

\ificcvfinal\fi

\begin{document}

\title{\vspace{-4mm}Learning Indoor Inverse Rendering with 3D Spatially-Varying Lighting}


\author{
  Zian Wang$^{1,2,3}$ \quad Jonah Philion$^{1,2,3}$ \quad Sanja Fidler$^{1,2,3}$ \quad Jan Kautz$^{1}$ \\
  NVIDIA\textsuperscript{1} \quad University of Toronto\textsuperscript{2} \quad Vector Institute\textsuperscript{3} \\
  {\tt\small \{zianw, jphilion, sfidler, jkautz\}@nvidia.com}
}

\maketitle
\ificcvfinal\thispagestyle{empty}\fi

\begin{abstract}
In this work, we address the problem of jointly estimating albedo, normals, depth and 3D spatially-varying lighting from a single image. 
Most existing methods formulate the task as image-to-image translation, ignoring the 3D properties of the scene. 
However, indoor scenes contain complex 3D light transport where a 2D representation is insufficient. 
In this paper, we propose a unified, learning-based inverse rendering framework that formulates 3D spatially-varying lighting. 
Inspired by classic volume rendering techniques, we propose a novel Volumetric Spherical Gaussian representation for lighting, which parameterizes the exitant radiance of the 3D scene surfaces on a voxel grid. 
We design a physics-based differentiable renderer that utilizes our 3D lighting representation, and formulates the energy-conserving image formation process that enables joint training of all intrinsic properties with the re-rendering constraint. Our model ensures physically correct predictions and avoids the need for ground-truth HDR lighting which is not easily accessible.
Experiments show that our method outperforms prior works both quantitatively and qualitatively, and is capable of producing photorealistic results for AR applications such as virtual object insertion even for highly specular objects. 
\end{abstract}

\vspace{-5mm}
\section{Introduction}
\label{sec:intro}
\vspace{-1mm}

The task of inverse rendering, originally proposed by Barrow and Tenenbaum~\cite{barrow1978recovering} in 1978, aims to reverse the rendering process by estimating reflectance, shape and lighting from a single image. Estimating these intrinsic properties enable downstream applications in augmented and mixed reality, such as realistic insertion of 3D objects into a given 2D image. 
Inverse rendering also facilitates semantic scene analysis such as object segmentation~\cite{baslamisli2018joint}. 

Given only observed pixel values, the problem of disambiguating reflectance, geometry and their complex interactions with lighting is challenging and ill-posed. Classic optimization-based methods leverage hand-crafted priors to constrain the ill-posed nature of the problem. However, these priors do not always hold for complex real world images and can lead to artifacts. 
Indoor scenes commonly encountered in AR applications are considered especially challenging due to complex 3D light transport that occurs indoors. 

\begin{figure}[t!]
\vspace{-1mm}
\centering
\footnotesize
\setlength{\tabcolsep}{0pt}
\begin{tabular}{ccccc}
\includegraphics[width=0.20\linewidth]{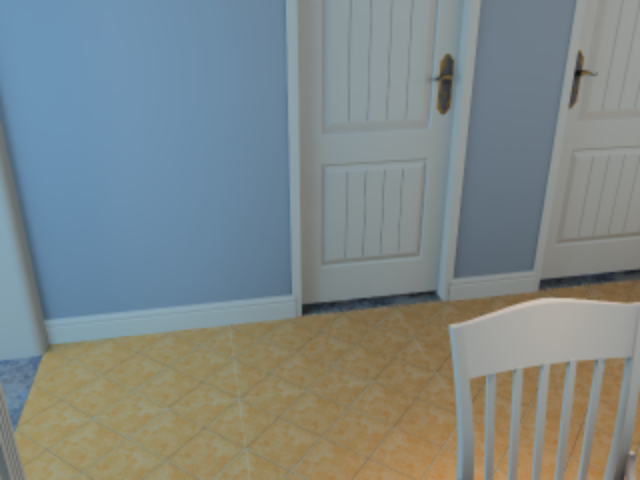} & \hspace{1mm}
\includegraphics[width=0.20\linewidth]{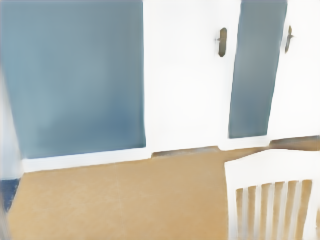} & 
\includegraphics[width=0.20\linewidth]{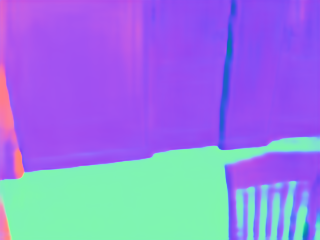} & 
\includegraphics[width=0.20\linewidth]{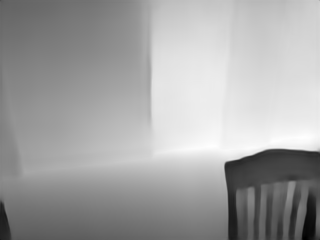} & \hspace{0.5mm}
\includegraphics[width=0.15\linewidth]{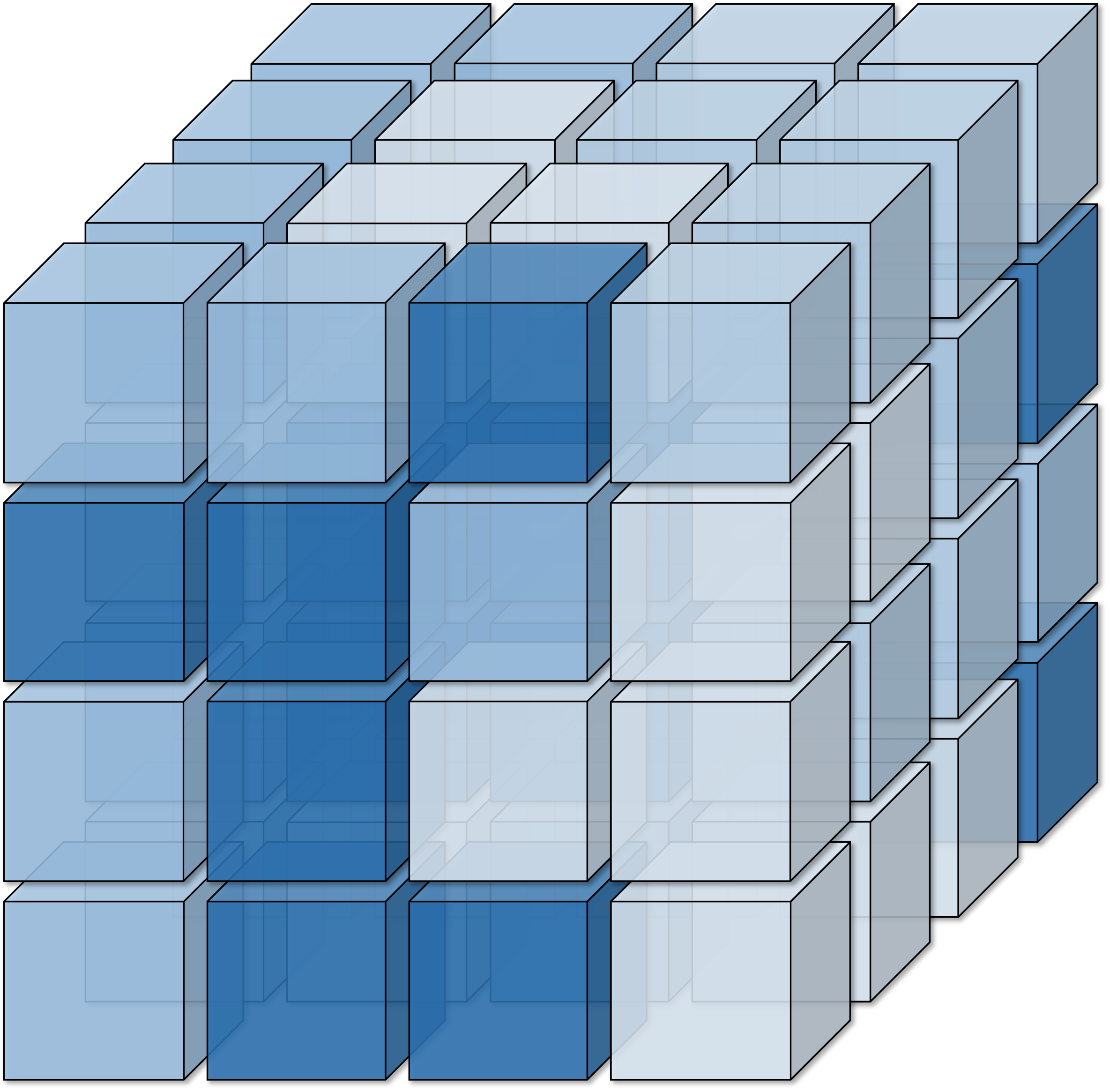} \\
(a) Input image & (b) Albedo  & (c) Normals  & (d) Depth  & (e) Lighting \\
\end{tabular}
\vspace{1mm}
\begin{tabular}{cc}
\includegraphics[width=0.49\linewidth,height=2.8cm,trim=0 50 0 0,clip]{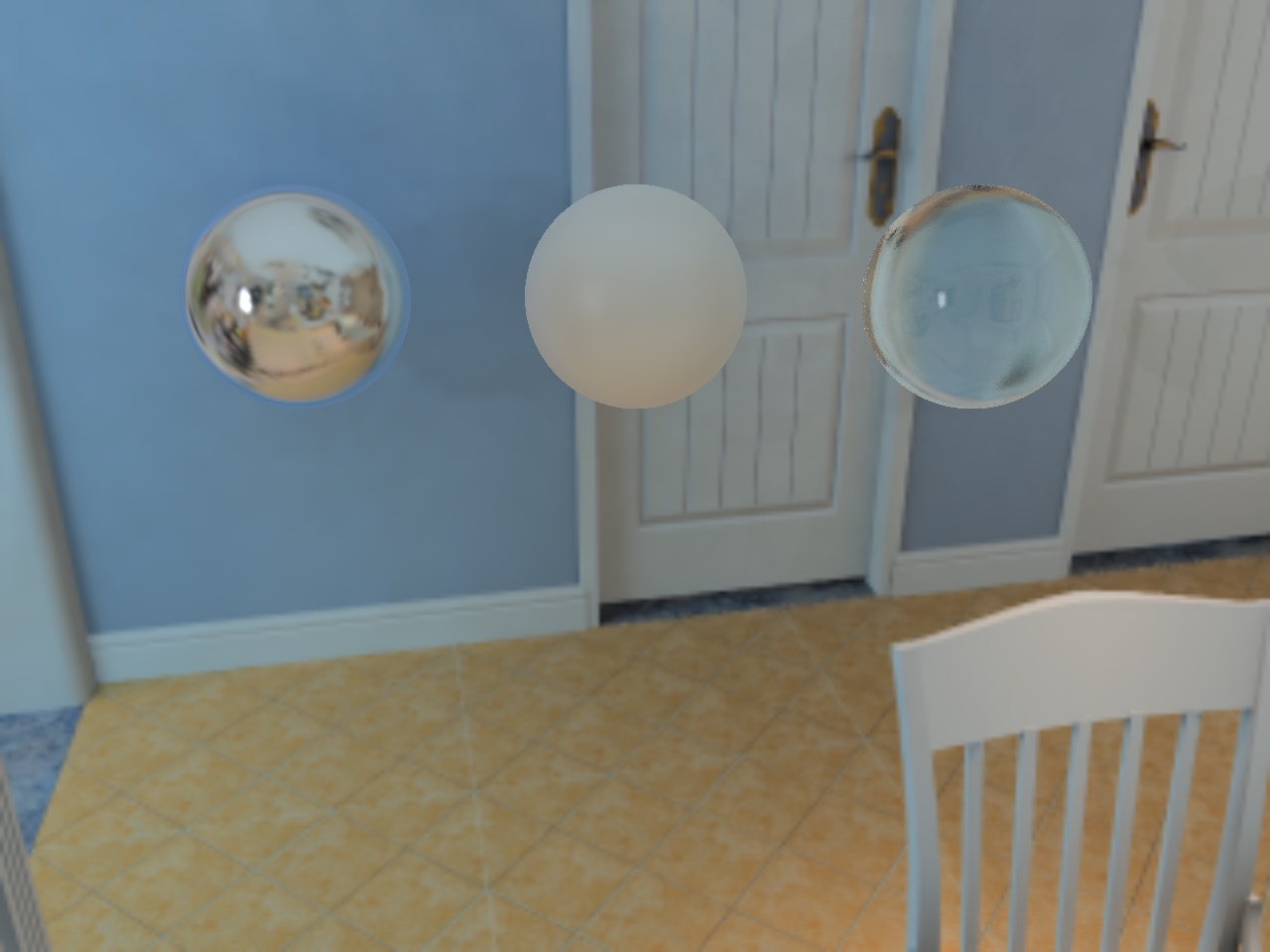} & \hspace{1mm}
\includegraphics[width=0.49\linewidth,height=2.8cm,trim=0 30 0 20,clip]{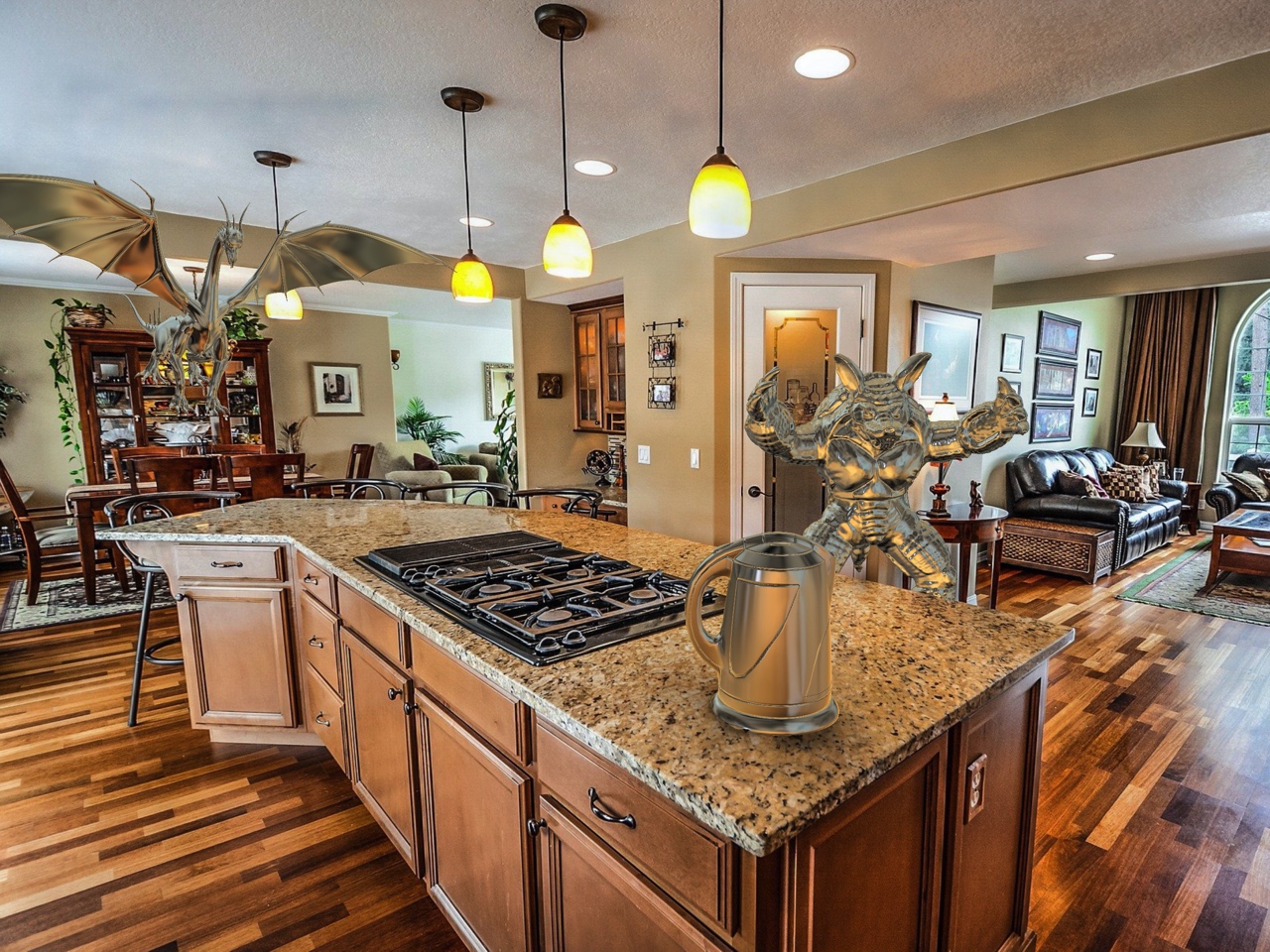} \\
\tabincell{c}{(f) Specular / Diffuse / Transparent \\ Sphere Insertion} & (g) Specular Object Insertion 
\end{tabular}
\vspace{-4mm}
\caption{\small\textbf{} From a single image, our model jointly estimates albedo, normals, depth, and the HDR lighting volume. Key to our method is inferring continuous HDR 3D spatially-varying lighting, which is critical in producing high quality virtual object insertion with realistic cast shadows and angular high-frequency details. }
\label{fig:teaser} 
\vspace{-5mm} 
\end{figure}

In this work, we address the problem of scene-level inverse rendering, focusing specifically on producing high dynamic range (HDR) 3D spatially-varying lighting with high-frequency details, as shown in Fig.~\ref{fig:teaser}. 
Estimating both HDR and 3D spatially-varying lighting is critical for photorealistic virtual object insertion; HDR enables realistic cast shadows and 3D spatially-varying lighting enables high-frequency details. We use the HDR lighting inferred by our model to insert highly specular objects and produce realistic cast shadows and high-frequency details, which were not possible in previous works \cite{garon2019fast,li2020inverse,neuralSengupta19,srinivasan2020lighthouse}. 

Existing learning-based methods usually exploit powerful 2D CNNs and formulate the inverse rendering problem as image-to-image translation. 
Lighting is usually represented with spherical lobes such as spherical Harmonics and spherical Gaussian \cite{barron2014shape,yu19inverserendernet}, and environment maps \cite{gardner2017learning,neuralSengupta19}, which ignores spatially-varying effects. 
Recent works attempt to predict 2D spatially-varying spherical lobes \cite{garon2019fast,li2020inverse}, but still lack one degree of freedom (depth) and compromise in terms of angular high-frequency effects. 
As a consequence, the 2D representation of the scene lighting is not sufficiently performant for many downstream applications.

In this paper, we propose a holistic inverse rendering framework for jointly estimating reflectance, shape and 3D spatially-varying lighting, by formulating the complete rendering process in an end-to-end trainable way with a 3D lighting representation. 
We propose a novel Volumetric Spherical Gaussian representation for lighting, which is a voxel representation for the scene surfaces. Spherical Gaussian parameters in each voxel control the emission direction and sharpness of the light source,  which captures view-dependent effects and can handle strong directional lighting. Since ground-truth for HDR lighting is not easily available, we design a raytracing based differentiable renderer that leverages our lighting representation and formulates the energy-conserving image formation process. We use the renderer to jointly train all intrinsic properties by enforcing the re-rendering constraint, ensuring that predictions are physically correct. 
To the best of our knowledge, our approach is the first to estimate a complete continuous light field function from a single image, including both HDR and high-frequency spatial and angular details, despite being trained with only LDR images. 

We experimentally show that our approach outperforms existing state-of-the-art inverse rendering and lighting estimation methods.
We demonstrate that our method  learns to produce complex lighting effects of real-world indoor scenes and better disambiguates intrinsic properties. 
Our lighting representation enables realistic cast shadows and angular high-frequency details and is therefore capable of producing significantly more realistic object insertion results for AR applications that were not possible previously, most importantly including the insertion of highly specular objects.

\vspace{-1mm}
\section{Related Work}

\vspace{-1mm}
\paragraph{Inverse Rendering. }
The task of inverse rendering dates back to Barrow and Tenenbaum~\cite{Barrow1978RECOVERINGIS}, with the goal of jointly estimating intrinsic properties of the scene, \ie reflectance, shape and lighting. 
Classic approaches usually tackle sub-tasks of inverse rendering, such as intrinsic image decomposition \cite{barrow1978recovering,bousseau2009user,mitberkeley,land1971lightness1,zhao2012closed} and shape from shading \cite{oxholm2012shape,zhang1999shape}. These methods primarily defined hand-crafted priors over the content of a scene and formulate the task as a per-image energy minimization problem. 
Recently, SIRFS~\cite{barron2014shape} proposed a statistical inference framework that jointly estimates the intrinsic properties. 
However, these methods rely on assumptions that are not always true for real scenes, leading to artifacts when applied on real-world images. 
The need to perform test-time optimization also raises the computational burden, limiting these approaches to offline applications. 

Recent works utilize 2D CNNs for learning data-driven priors from sparse human reflectance annotations~\cite{bell2014intrinsic}, calibrated multi-view and multi-illumination data~\cite{li2018learningWatching,yu19inverserendernet}, and most commonly, synthetic data~\cite{boss2020two,li2020inverse,li2018cgintrinsics,li2018learning,neuralSengupta19} that comes with dense ground-truth labels. 
Among these works,~\cite{boss2020two,li2018learning} are limited to single object inputs and do not address complex light transport.
NIR~\cite{neuralSengupta19} and Li \etal \cite{li2020inverse} are most similar to our work, and address general indoor scenes. Both formulate the task as image-to-image translation and train on synthetic datasets. 
NIR employs an environment map to represent lighting and introduces a non-interpretable neural renderer to account for spatially-varying lighting effects.
Li \etal  predicts a spherical Gaussian lighting for each pixel location to get 2D spatially-varying lighting, but it still lacks one degree of freedom and sacrifices angular frequency at each pixel location. 
In our work, we tackle a more challenging task of estimating 3D lighting in a holistic inverse rendering framework, and learn to disentangle complex lighting effects with a physics-based representation.

\vspace{-4mm}
\paragraph{Lighting Estimation} is a sub-task of inverse rendering.
Most existing works tackle simplified problem settings and ignore spatially-varying effects, such as outdoor scenes \cite{hold2019deep,hold2017deep,zhang2019all} and objects \cite{boss2020two,li2018learning,wei2020object}. 
Prior works on indoor lighting estimation explored lighting representations such as environment maps \cite{gardner2017learning,legendre2019deeplight,neuralSengupta19}, per-pixel spherical lobes \cite{garon2019fast,li2020inverse,zhao2020pointar} and light source parameters~\cite{gardner2019deep}. 
However, these methods either do not account for spatially-varying effects or do not preserve high-frequency details. 
Recent works explore 3D spatially-varying lighting~\cite{song2019neural,srinivasan2020lighthouse}. 
Neural Illumination~\cite{song2019neural} predicts an environment map with 2D CNNs given each queried 3D location, but it suffers from spatial instability. 
Lighthouse~\cite{srinivasan2020lighthouse} tackles this problem from the perspective of view synthesis and confirms the necessity for a 3D lighting representation. However, it does not address HDR information, and there is no guarantee that the inpainted lighting is physically correct. 
In this work, we leverage a holistic inverse rendering framework to enable physically correct HDR lighting prediction, whereby we only train on the LDR ground-truth data as in Lighthouse.

\vspace{-4mm}
\paragraph{Neural Scene Representations. } 
Efficient 3D representation is a rapid area of research, such as voxels~\cite{lombardi2019neural,sitzmann2019deepvoxels}, point clouds~\cite{aliev2019neural,meshry2019neural} and implicit functions \cite{mescheder2019occupancy,mildenhall2020nerf,sitzmann2019scene,takikawa2021nglod}. 
DeepVoxels~\cite{sitzmann2019deepvoxels} and NeuralVolumes~\cite{lombardi2019neural} use neural networks to predict voxel representations of scenes, where each voxel grid contains neural features or RGB$\alpha$ values respectively. 
Zhou \etal \cite{zhou2018stereo} proposed to use multi-plane images, which is an voxelized representation for the camera frustum. 
When combined with volume rendering, these works enable predicting 3D properties but require only 2D supervision. 
Recent methods also show promising results with neural implicit functions~\cite{mescheder2019occupancy,mildenhall2020nerf,sitzmann2019scene}, which represent the scene as a continuous volumetric function and approximate with a neural network. 
Our proposed Volumetric Spherical Gaussian draws inspiration from 3D scene representations and augments the RGB$\alpha$ representation with view-dependent effects to better handle directional lighting.


\begin{figure*}[t!]
\centering
\vspace{-1mm}
\begin{minipage}{0.68\linewidth}
\vspace{-4mm}
\includegraphics[width=0.99\linewidth]{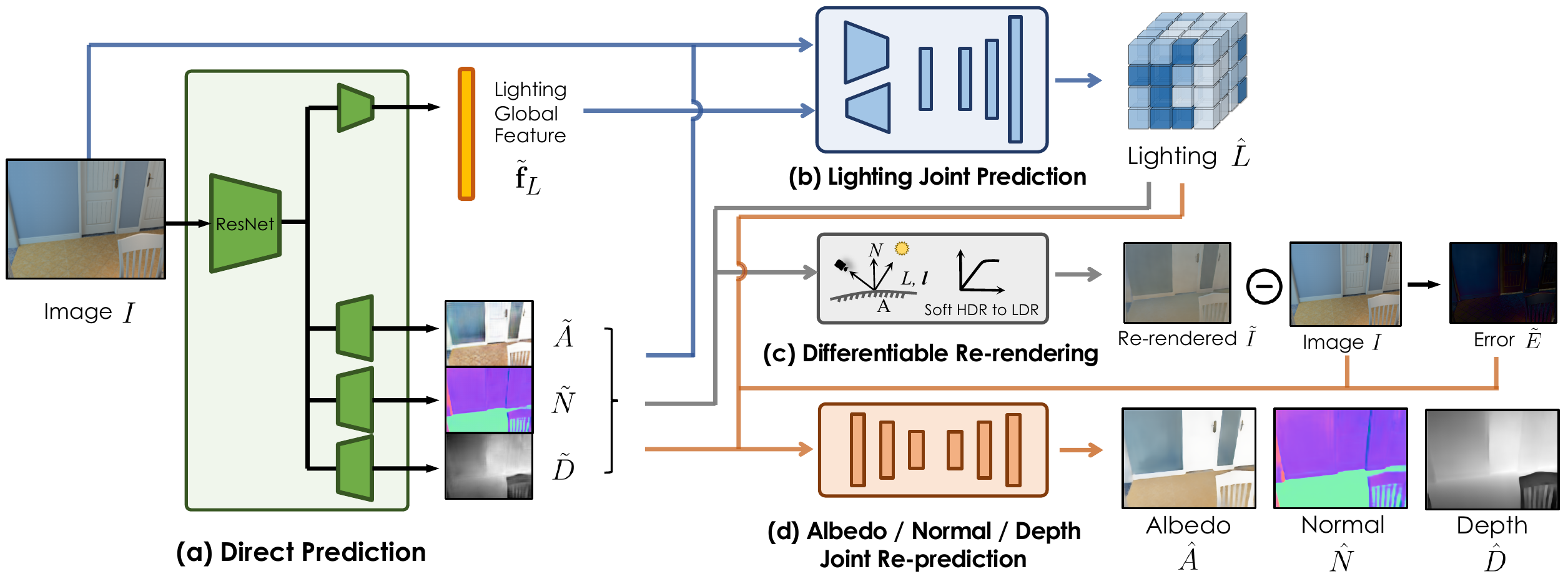} 
\end{minipage}
\begin{minipage}{0.015\linewidth}
\end{minipage}
\begin{minipage}{0.30\linewidth}
\caption{\small\textbf{Model overview.} Our model consists of 4 submodules (a-d). 
Direct Prediction Module (a) takes a single image as input and jointly predicts initial guess of intrinsic properties. 
Lighting Joint Prediction Module (b) consumes the initial prediction and predicts a 3D lighting volume. 
With Differentiable Re-rendering Module (c) re-renders the input image by raytracing, 
Joint Re-prediction Module (d) finally jointly refines the initial prediction. 
}
\label{fig:model} 
\end{minipage}
\vspace{-7mm}
\end{figure*}

\vspace{-1mm}
\section{Lighting Representation} 
\label{sec:vsg}
\vspace{-1mm}

To invert the rendering process, we require a flexible, structured representation of 3D spatially-varying lighting. 
Ideally, a model of the light field should capture variation in radiance due to changes in both spatial location and viewing angle. 
Prior works \cite{garon2019fast,li2020inverse} represent lighting with the radiance \emph{incident} at the visible surface, making the direct extension to 3D intractable. 
To address this issue, we propose to use Volumetric Spherical Gaussian (VSG) to represent the surface radiance \emph{exitant} from the full scene, including both visible surfaces and surfaces outside the FoV. Illumination at any spatial location and viewing angle can then be rendered using standard volume rendering techniques. 

VSG is a voxel-based representation of a scene.
We assign an opacity $\alpha \in [0, 1]$ to each voxel, as well as a set of spherical Gaussian parameters $\mathbf{c} \in \mathbb{R}^3$, $\bm{\mu} \in \mathbb{R}^3$, $\sigma \in \mathbb{R}_+$ such that the radiance at a viewing angle $\bm{v} \in \mathbb{R}^3$ is defined:\\[-2mm]
\begin{equation}
	G(\bm{v}; \bm{c}, \bm{\mu}, \sigma) = \bm{c}e^{-(1 - \bm{v} \cdot \bm{\mu}) / \sigma^2}
	\vspace{-0.5mm}
\end{equation}
Intuitively, every voxel is a light source, where $\bm{c}$ represents HDR RGB intensity, $\bm{\mu}$ is the lobe axis and $1/\sigma^2$ indicates sharpness. 
For a voxel grid of size $X \times Y \times Z$, VSG represents the lighting as an 8-channel tensor $L \in \mathbb{R}^{8 \times X \times Y \times Z}$.

To calculate incident radiance for a point $\bm{p} \in \mathbb{R}^3$ with direction $\bm{l} \in \mathbb{R}^3$ in a VSG $L$, we select $N$ equi-spaced points along the ray.
We then calculate the radiance $\mathcal{R}(\bm{p}, \bm{l}, L) \in \mathbb{R}^3$ using alpha compositing
\begin{equation}
	\mathcal{R}(\bm{p}, \bm{l}, L) = \sum_{k=1}^{N} \prod_{i=1}^{k-1} (1-\alpha_{i}) \alpha_{k} G(-\bm{l}; \bm{c}_k, \bm{\mu}_k, \sigma_k). 
	\label{eq:alphacomp}
\end{equation}
where $\bm{c}_k, \bm{\mu}_k, \sigma_k$ and $\alpha_k$ are determined by indexing into the lighting volume $L$. 
With $\alpha$-channel indicating the opacity of voxels, we can also render ``depth'' with Eq.~\ref{eq:alphacomp} by replacing the spherical Gaussian with the voxel depth values. 

Compared to recent works that use an RGB$\alpha$ volume to render appearance~\cite{mildenhall2020nerf,srinivasan2020lighthouse}, the VSG lighting representation additionally controls the emission direction and sharpness of light sources, and thus can capture view-dependent effects and handle strong directional lighting. Note that for $\sigma \gg 1$, our VSG reduces to an RGB$\alpha$ representation.

\vspace{-1mm} 
\section{Method} 
\label{sec:method} 
\vspace{-1mm} 

Our monocular inverse rendering model jointly estimates albedo, normals, depth and a 3D Volumetric Spherical Gaussian lighting representation. 
To jointly predict intrinsic properties and 3D VSG lighting from a single image, we split our pipeline into four submodules, with three neural network modules and one differentiable rendering module. The overall architecture is shown in Fig.~\ref{fig:model}. 

First, the Direct Prediction Module makes an initial prediction of the intrinsic properties and extracts a global lighting feature from an input image. 
The Joint Prediction Module lifts these properties into 3D and jointly predicts a lighting volume. 
Then, the Differentiable Re-rendering Module re-renders the input image using the current predicted intrinsics and lighting. 
Finally, the Joint Re-prediction Module conditions on the current prediction and re-rendering error, and jointly refines the initial predictions.

We describe the architecture of each submodule in Sec.~\ref{ssec:arch} and present our training scheme in Sec.~\ref{ssec:training}. 

\subsection{Model Design} 
\label{ssec:arch}

\vspace{-1mm}
\paragraph{Direct Prediction Module. } 
The purpose of the Direct Prediction Module is to make an initial prediction of the albedo, normal, depth and a global feature which encodes lighting information, given a single image as input. 

The backbone of the Direct Prediction Module is a multi-branch ResNet \cite{resnet} $h_\text{DP}$ that consumes a single linearized RGB image $I \in \mathbb{R}^{3 \times H \times W}$ and predicts albedo $\tilde{A} \in \mathbb{R}^{3 \times H \times W}$, surface normals $\tilde{N} \in \mathbb{R}^{3 \times H \times W}$, depth $\tilde{D} \in \mathbb{R}^{H \times W}$ and a global feature vector $\tilde{\mathbf{f}}_L \in \mathbb{R}^C$ that is used by the downstream Lighting Joint Prediction Module as a ``Lighting Global Feature''
\begin{equation} 
	\tilde{A}, \tilde{N}, \tilde{D}, \tilde{\mathbf{f}}_L = h_\text{DP}(I; \Theta_\text{DP}).
\end{equation} 

\begin{figure}[t]
\centering
\begin{minipage}{1.0\linewidth}
\centering
\includegraphics[width=0.99\linewidth]{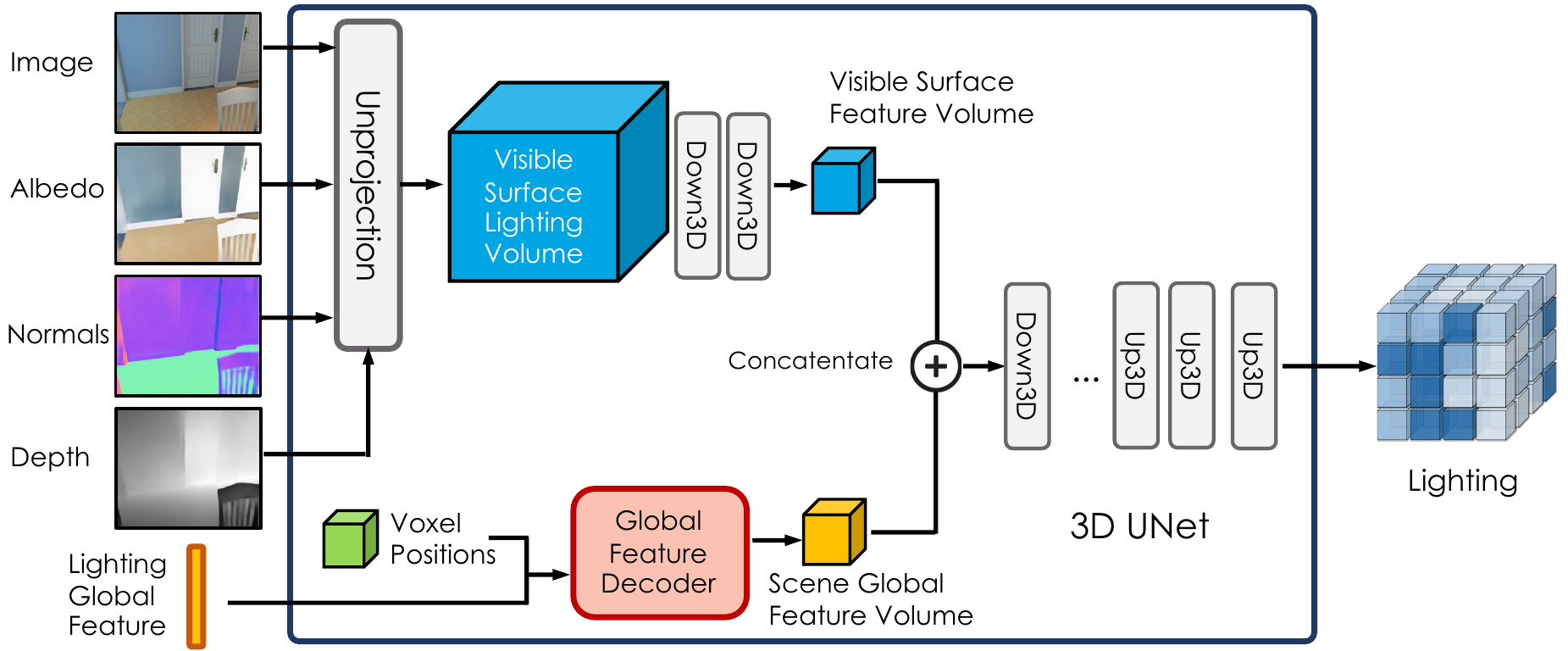} 
\vspace{-2mm}
\caption{\small\textbf{Architecture of Lighting Joint Prediction Module.} 
We fuse the unprojected visible FoV information (top) and global scene information (bottom), and process them with a 3D UNet. The output is the Volumetric Spherical Gaussian lighting.}
\label{fig:LDN} 
\end{minipage}
\vspace{-5mm} 
\end{figure}

\vspace{-6.9mm}
\paragraph{Lighting Joint Prediction Module. } 
Unlike albedo, normals, and depth, our lighting representation as defined in Sec.~\ref{sec:vsg} is volumetric: $\hat{L} \in \mathbb{R}^{8 \times X \times Y \times Z}$. Our lighting decoding network  is shown in Fig.~\ref{fig:LDN} and described below. 

We extract features to predict the lighting volume from two different sources. The first source is the global feature vector $\tilde{\mathbf{f}}_L$. 
We use an MLP decoder $h_\text{GFD}$ to map the global feature into a Scene Global Feature Volume. 
Let $(x, y, z)$ be the center coordinates of a given voxel. 
The feature at the corresponding voxel is computed as:
$$\bm{z}_g = h_\text{GFD}(x, y, z, \tilde{\mathbf{f}}_L).$$ 
Compared to a sequence of 3D transpose convolution which can achieve similar functionality, this MLP module is more flexible and can naturally extend to multi-view input. 
We refer to Appendix for more implementation details. 

The other sources of features for the lighting volume are the properties within the visible FoV, including input image $I$, predicted albedo $\tilde{A}$, normal $\tilde{N}$, and depth $\tilde{D}$. 
We unproject this visible FoV information into a Visible Surface Lighting Volume and process it with a 3D UNet. 
Given camera intrinsics, let $(u_p,v_p)$ be the projection of the center point of the voxel onto the input image with depth $d_p$, let $\tilde{D}_p,\tilde{N}_p,\tilde{A}_p$ be the depth, normal, and albedo predicted by $h_\text{DP}$ at the pixel $(u_p, v_p)$, and let $I_p$ be the RGB values of the input image at that pixel. For each voxel, we define its ``local'' features as:\\[-3mm]
\begin{equation}
\label{eq:locallight}
\begin{aligned}
    \bm{z}_l &= kI_p, k\tilde{N}_p, k\tilde{A}_p
\end{aligned}
\end{equation}
where $k=e^{-\frac{(d_p - \tilde{D}_p)^2}{2\sigma_\text{d}^2}}$ is the Gaussian distance of depth between the voxel and the corresponding pixel, 
and $\sigma_d$ is a hyper-parameter with units of length that we set to $0.15$ meters. 
Intuitively, the factor $k$ zeros out local features for voxels that are far from 2D surface manifold as determined by the depth output from $h_\text{DP}$. 

We fuse the global features $\mathbf{z}_g \in \mathbb{R}^{C \times X_g \times Y_g \times Z_g}$ and local features of the visible FoV $\mathbf{z}_l \in \mathbb{R}^{9 \times X \times Y \times Z}$, and process them through a 3D CNN $h_\text{JP}$: 
\begin{equation} 
	\hat{L} = h_\text{JP}(\mathbf{z}_g, \mathbf{z}_l; \Theta_\text{JP}),\quad \hat{L} \in \mathbb{R}^{8 \times X \times Y \times Z}
\end{equation} 
where $\hat{L}$ is the output VSG containing HDR intensity.

\vspace{-4mm}
\paragraph{Differentiable Re-rendering Module.}
A valuable source of supervision and inductive bias in inverse rendering comes from the fact that the input image should be retrieved if the predicted geometry and reflectance is re-rendered. 

We use the Lambertian reflectance model for re-rendering. For each pixel, let $\bm{p} = (x_p, y_p, \tilde{D}_p)$ be the 3D location of the pixel $p$ with predicted depth $\tilde{D}_p$. 
To compute lighting at each pixel, we choose $K$ equi-angular lighting directions $\{\bm{l}\}_K$ on the upper hemisphere with Fibonacci lattice \cite{gonzalez2010measurement}. For each lighting direction $\bm{l}$, we query the lighting volume and compute the radiance along the ray $\mathcal{R}(\bm{p}, \bm{l}, \hat{L})$ with Eq.~\ref{eq:alphacomp}. 

Let $\tilde{A}_p$ and $\tilde{N}_p$ be albedo and normal predicted in pixel $p$.
We render LDR RGB values following  Lambertian model:
\vspace{-1mm}
\begin{align} 
\label{eq:lambertian}
    \tilde{I}_p = \varphi \Big(\sum_{\bm{l} \in \{ \bm{l}\}_K} \frac{\tilde{A}_p}{\pi} \odot \mathcal{R}(\bm{p}, \bm{l}, \hat{L}) \max(\bm{l} \cdot \tilde{N}_p, 0) \Delta\Omega \Big) 
\end{align} 
where $\odot$ is an element-wise product, $\frac{1}{\pi}$ is the energy conservation ratio and $\Delta\Omega$ is the differential solid angle. 
Here, $\varphi$ is the function to clip the HDR lighting intensity values to LDR values within $[0, 1]$. 
To make this process differentiable, we use a soft clipping with the exponential function:
\vspace{-1mm}
\begin{equation}
    \varphi (x) = \left\{
    \begin{aligned}
        &x                                             & \text{if} \ x \le \tau \\
        &1-(1-\tau) e^{-\frac{x-\tau}{1-\tau}}     & \text{if} \ x > \tau 
    \end{aligned}
    \right. 
    \label{eq:hdr2ldr}
\end{equation}
We use $\tau=0.9$ in our model.

\vspace{-4mm}
\paragraph{Joint Re-prediction Module. } 
Reflectance, shape and lighting are inherently correlated through the rendering process. 
To refine albedo, normals, and depth, we use a fully-convolutional network that takes as input the initial prediction $\tilde{A}$, $\tilde{N}$, $\tilde{D}$, the rerendering error $\tilde{E}=I-\tilde{I}$, and input image $I$.  
To incorporate the predicted lighting $\hat{L}$, we also concatenate shading $\tilde{S}$, which indicates how albedo changes affect the re-rendered image, and the Jacobian of shading with respect to the normals $\frac{\partial \tilde{S}}{\partial \tilde{N}}$, which indicates how the normals' change affect the output shading. 
At each pixel $p$, $\tilde{S}$ and $\frac{\partial \tilde{S}}{\partial \tilde{N}}$ have values given by the analytic formulas 
\vspace{-1mm}
\begin{equation}
\label{eq:JR_lighting}
    \begin{aligned}
    \tilde{S}_p &= \sum_{\bm{l} \in \{ \bm{l}\}_K} \mathcal{R}(\bm{p}, \bm{l}, \hat{L}) \max(\bm{l} \cdot \tilde{N}_p, 0) \Delta\Omega\\
    \frac{\partial \tilde{S}}{\partial \tilde{N}}_p &= 
    \sum_{\bm{l} \in \{ \bm{l}\}_K} \mathbbm{1}_{\bm{l} \cdot \tilde{N}_p > 0} \mathcal{R}(\bm{p}, \bm{l}, \hat{L}) \otimes \bm{l} \Delta\Omega
    \end{aligned}
\end{equation} 
where $\otimes$ is the outer product. 
Note that much of the computation required for shading can be cached during Lambertian rendering given by Eq.~\ref{eq:lambertian}. The fully-convolutional network predicts an updated albedo $\hat{A}$, normals $\hat{N}$, and depth $\hat{D}$
\vspace{-1mm}
\begin{equation}
    \hat{A}, \hat{N}, \hat{D} = h_{\text{JR}}(I, \tilde{E}, \tilde{A}, \tilde{N}, \tilde{D}, \tilde{S}, \frac{\partial \tilde{S}}{\partial \tilde{N}}).
\end{equation}

\subsection{Training} 
\label{ssec:training}
We train our model on synthetic data with groundtruth $\{I, A, N, D, \{I_\text{nv}, P_\text{nv}\}_N\}$, where $A, N, D$ denote albedo, normals, depth, and $\{I_\text{nv}, P_\text{nv}\}_N$ are LDR panoramic images and camera poses from $N$ novel views. 
With a volumetric lighting representation, we not only eliminate the need for the densely rendered spherical lobe lighting GT used in past works~\cite{li2020inverse,li2020openrooms}, but also improve angular frequency. 

The loss for training comes from two parts: (1) \emph{direct supervision}, which directly enforces consistency with the synthetic groundtruth, and (2) \emph{re-rendering loss} that encourages the re-rendered image to recover the input image. 

\vspace{-4mm}
\paragraph{Direct Supervision for Reflectance and Shape. } 
We use L2 loss for albedo. Since albedo is usually piecewise constant, we additionally penalize the gradient of albedo where groundtruth is locally constant. We define 
\begin{equation}
    \mathcal{L}_{\text{albedo}} = ||A - \hat{A}||_2^2 + \lambda_\text{local}||\nabla \hat{A} \odot M_\text{local}||_1
\end{equation}
where $M_{\mathrm{local}} \in \mathbb{R}^{H \times W}$ is a mask indicating regions in the ground-truth albedo where the albedo is constant. 

For normals, the network output is normalized and we use L1 angular error as supervision:
\vspace{-2mm} 
\begin{equation} 
    \mathcal{L}_{\text{normal}} = ||\cos^{-1}(N \cdot \frac{\hat{N}}{|\hat{N}|})||_1.
\vspace{-2mm} 
\end{equation}

Because depth is high dynamic range, we follow \cite{li2020inverse} and use log-encoded L2 loss. 
We also use scale invariant L2 loss to encourage relative consistency due to the inherent scale ambiguity of depth: 
\vspace{-2mm} 
\begin{equation}
\begin{aligned}
    \mathcal{L}_{\text{depth}} &= ||\log(D+1) - \log(\hat{D}+1)||_2^2\\ 
    &+ \lambda_\text{si}||D - c_\text{si}\hat{D}||_2^2 
\end{aligned}
\vspace{-2mm} 
\end{equation}
where $c_\text{si} = \arg\min_{c} ||D - c\hat{D}||_2^2 = \frac{\sum_{p} \hat{D}_p \cdot D_p}{\sum_p\hat{D}_p \cdot \hat{D}_p}$ is a scale factor computed on-the-fly for each image. 

\vspace{-4mm}
\paragraph{Direct Supervision for Lighting. } 
Recall that the pixel values in LDR images reflect the HDR radiance along the corresponding camera ray, after intensity clipping. 
Thus, we can use photometric loss of LDR panoramic images $I_\text{nv}$ to supervise the LDR part of the predicted lighting $\hat{L}$. 

For each pixel $p$, we use the camera pose $P_\text{nv}$ and camera intrinsics to compute the corresponding camera ray starting from the camera center $\bm{c}$ in the direction $\bm{r}$.
To render the novel view $\hat{I}_{\text{nv}, p}$ using the predicted lighting volume $\hat{L}$, 
we compute HDR radiance with Eq.~\ref{eq:alphacomp}, and convert to LDR using a soft clipping function $\varphi$ defined in Eq.~\ref{eq:hdr2ldr}:
\begin{equation}
    \hat{I}_{\text{nv}, p} = \varphi\big( \mathcal{R}(\bm{c}, \bm{r}, \hat{L}) \big)
\end{equation}
We enforce this rendered novel view $\hat{I}_\text{nv}$ to be consistent with groundtruth $I_\text{nv}$ using L2 loss. 
To encourage realistic details, we also use the adversarial loss $\mathcal{L}_{\text{adv}}$ with a discriminator $\mathcal{D}$. 
\begin{align}
    \label{eq:loss_L}
    \mathcal{L}_{\text{light}} &= \mathcal{L}_{\text{nv}} + \lambda_\text{adv} \mathcal{L}_{\text{adv}} \\
    &= ||I_\text{nv} - \hat{I}_\text{nv}||_2^2 - \lambda_\text{adv} \mathcal{D}(\hat{I}_\text{nv}) \notag
\end{align}
The loss for the discriminator $\mathcal{D}$ is
\begin{align}
    \mathcal{L}_{\text{D}} =& \max(0, 1-\mathcal{D}(I_\text{nv})) + \max(0, 1+\mathcal{D}(\hat{I}_\text{nv})). 
\end{align}

Another source of supervision for lighting is consistency with visible FoV, \ie the image $I$ and depth $D$. 
We define $\mathcal{L}_\text{visible}$ as the L2 loss between $I, D$ and the rendered perspective RGB image and depth ($\alpha$-channel) from the lighting volume using Eq.~\ref{eq:alphacomp}. 
Since the surface is sparse in the scene, we also encourage the $\alpha$-channel of the lighting volume to be either $0$ or $1$ with a regularization loss $\mathcal{L}_\text{reg}=-\alpha \log(\alpha)$. 
The training signal is fully differentiable and backpropagated to supervise the predicted VSG lighting parameters.

\begin{table*}[t]
\vspace{-1mm}
\centering
\begin{minipage}[t]{0.38\linewidth}
\centering
\resizebox{0.99\textwidth}{!}{
\begin{tabular}{|l|p{1.5cm}<{\centering}|p{2cm}<{\centering}|p{1.5cm}<{\centering}|}
\hline
Method & \tabincell{c}{Albedo \\ si-MSE}  & \tabincell{c}{Normal \\ Angular Error}  & \tabincell{c}{Depth \\ si-MSE}  \\
\hline
SIRFS \cite{barron2014shape}    & $0.0453$      & $56.75^{\circ} $      & -  \\
NIR \cite{neuralSengupta19}     & $0.0188$      & $20.35^{\circ}$       & -  \\
Ours (w/o JR)                   & $0.0190$      & $19.09^{\circ}$       & $0.217$  \\
Ours (JR w/o lighting)          & $0.0177$      & $18.63^{\circ}$       & $0.189$  \\
Ours                            & $\bm{0.0175}$ & $\bm{18.40^{\circ}}$  & $\bm{0.181}$  \\
\hline
\end{tabular}
} 
\vspace{-2.5mm} 
\caption{\small Evaluation of albedo, normals and depth on InteriorNet dataset. } 
\vspace{-3.5mm}
\label{table:interiornet_AND} 
\end{minipage} \hspace{2mm}
\begin{minipage}[t]{0.35\linewidth}
\vspace{-12.5mm}
\centering
\resizebox{0.9\linewidth}{!}{
\begin{tabular}{|l|c|}
\hline
Method & PSNR (dB) \\
\hline
NIR \cite{neuralSengupta19}     &  $15.39$  \\
Lighthouse$^*$ \cite{srinivasan2020lighthouse} &  $17.29$  \\
Ours ($\mathcal{L}_{\text{albedo}}, \mathcal{L}_{\text{normal}}, \mathcal{L}_{\text{depth}}, \mathcal{L}_{\text{light}}$ only)                     &  ${16.43}$  \\
+$\mathcal{L}_{\text{visible}}$         &  ${17.06}$  \\
+$\mathcal{L}_{\text{reg}}$             &  ${17.33}$  \\
+$\mathcal{L}_{\text{rerender}}$        &  $\bm{17.37}$  \\
Ours (w/o SG)                   &  ${16.94}$  \\
\hline
\end{tabular}
}
\vspace{-3mm}
\caption{\small Evaluation of lighting on InteriorNet dataset. * indicates use of a stereo pair as input. }
\vspace{-5mm}
\label{tab:interiornet_L}
\end{minipage} \hspace{2mm}
\begin{minipage}[t]{0.21\linewidth}
\vspace{-9mm}
\centering
\resizebox{0.9\textwidth}{!}{
\begin{tabular}{|l|c|}
\hline
Method & WHDR \\
\hline
SIRFS \cite{barron2014shape}    & $31.4$ \\
NIR \cite{neuralSengupta19}     & $18.5$ \\
Ours (w/o JR)                   & $18.7$ \\
Ours                            & $\bm{18.2}$ \\
\hline
\end{tabular}
} 
\vspace{-1mm}
\caption{\small Evaluation of albedo on IIW dataset. } 
\vspace{-3.5mm}
\label{table:iiw}
\end{minipage}
\vspace{-2mm}
\end{table*}

\begin{table}[t]
\centering
\resizebox{0.44\textwidth}{!}{
\begin{tabular}{|l|p{3cm}<{\centering}|p{3cm}<{\centering}|}
\hline
Method & \tabincell{c}{Normal Angular Error}  & \tabincell{c}{Depth si-MSE}  \\
\hline
NIR \cite{neuralSengupta19} & $23.94^{\circ}$   & $0.3216$  \\       
Ours (w/o JR)   & $23.89^{\circ}$       & $0.3196$  \\
Ours            & $\bm{22.95^{\circ}}$  & $\bm{0.2827}$  \\
\hline
\end{tabular}
} 
\vspace{-3mm} 
\caption{\small Evaluation of normals and depth on NYUv2 dataset. 
} 
\label{table:nyuv2} 
\vspace{-3mm}
\end{table}

\begin{table}[t]
\centering
\resizebox{0.44\textwidth}{!}{
\begin{tabular}{|l|c|c|}
\hline
Re-rendering MSE ($\times 10^{-2}$) & \tabincell{c}{InteriorNet \cite{InteriorNet18}}  & \tabincell{c}{NYUv2 \cite{Silberman:ECCV12}}  \\
\hline
NIR (env. map only) \cite{neuralSengupta19} & $2.36$      & $4.02$  \\       
NIR \cite{neuralSengupta19}                 & $0.99$      & $2.61$  \\       
Ours (w/o Re-render Loss)                   & $2.18$      & $5.26$  \\
Ours (w/o SG)                               & $1.41$      & $2.72$  \\
Ours                                        & $\bm{0.89}$ & $2.33$  \\
Ours (w/ real-world tuning)                 & $0.92$      & $\bm{1.98}$  \\
\hline
\end{tabular}
} 
\vspace{-3mm} 
\caption{\small Quantitative results of re-rendering error. 
} 
\label{table:rerender} 
\vspace{-4.5mm}
\end{table}

\vspace{-4mm}
\paragraph{Re-rendering Loss. }
We use the predicted albedo, normals and lighting to reconstruct an image $\hat{I}$ and enforce its consistency with the original input image $I$. 
Specifically, we compute Lambertian re-rendered image in the energy conserved form defined in Eq.~\ref{eq:lambertian}, 
and use L2 loss as the re-rendering loss $\mathcal{L}_{\text{rerender}} = ||I - \hat{I}||_2^2$. 

The re-rendering loss encourages joint reasoning of albedo, geometry and lighting, and can be used for self-supervised training on real world images as discussed in prior works~\cite{neuralSengupta19}. 
In our formulation, the re-rendering loss significantly improves lighting prediction by enforcing the model to learn physically correct lighting and to recover HDR information. 
This benefit comes from the formulation of the energy-conserving image formation process. 
Any physically incorrect lighting prediction, such as a LDR prediction or a uniform lighting, will lead to an error in the re-rendered image. 
In Eq.~\ref{eq:loss_L}, we only supervise LDR lighting appearance with LDR images. With the complementary re-rendering loss, our model automatically learns to recover HDR lighting even when trained only with LDR images.

\vspace{-4mm}
\paragraph{Training Scheme. } 
Our model is end-to-end trainable. We adopt a progressive training scheme to ensure the model components act as expected. 
We first pretrain albedo, normal and depth branches of our Direct Prediction Module. This is because our Lighting Joint Prediction Module (Fig.~\ref{fig:LDN}) depends on these properties, and pretraining these branches can make sure they produce reasonable values. 
Then we jointly train the Direct Prediction Module and Lighting Joint Prediction Module with the multi-task loss
\begin{align} 
    \mathcal{L} =& \lambda_A \mathcal{L}_{\text{albedo}} + \lambda_N \mathcal{L}_{\text{normal}} + \lambda_D \mathcal{L}_{\text{depth}} + \lambda_L \mathcal{L}_{\text{light}} \notag \\
    &+ \lambda_\text{visible} \mathcal{L}_{\text{visible}} + \lambda_\text{reg} \mathcal{L}_{\text{reg}} + \lambda_\text{rerender} \mathcal{L}_{\text{rerender}} .
\end{align} 
After the first two submodules are trained, we freeze their weights and train the Joint Re-prediction Module for albedo, normals and depth. 
Finally, we jointly finetune all three modules end-to-end with a multi-task loss on both the the joint prediction $\hat{A}, \hat{N}, \hat{D}, \hat{L}$ and intermediate output $\tilde{A}, \tilde{N}, \tilde{D}$.

\vspace{-1mm}
\section{Experiments}
\label{sec:results}
\vspace{-1mm}

We compare our method with prior methods both qualitatively and quantitatively, and validate the effectiveness of our unified inverse rendering framework. 
We also compare against prior methods on lighting estimation and showcase application on virtual object insertion, demonstrating our method’s ability to create high-quality insertion results.

\subsection{Experiment Settings} 

\vspace{-1mm}
\paragraph{Implementation Details. } 
The resolution of initialized visible surface volume and predicted VSG lighting is $128^3$. 
Despite using a $128^3$ volume, our model is much lighter than Lighthouse \cite{srinivasan2020lighthouse} which contains six 3D UNet subnetworks. 
With a batch size of 1, our model consumes 7.5G GPU memory compared to Lighthouse's 15G during training. 
For Differentiable Re-rendering Module, we sample $K=50$ rays per pixel and $N=128$ points per ray. 
For each pixel, we share the rays of its 8-neighbors to get $K^{'}=450$ rays per pixel. 
The re-rendering is implemented considering parallelism. Resolution of re-rendering is 60x80 during training. 
Inference times of Direct Prediction, Lighting Joint Prediction, Differentiable Re-rendering and Joint Re-prediction Modules are 20ms, 130ms, 140ms, 12ms respectively, clocked on a TITAN V GPU.

\vspace{-4mm}
\paragraph{Training Data.}
We train our model on the InteriorNet dataset~\cite{InteriorNet18}, which contains realistic renderings of camera sequences in diverse indoor scenes. Each camera sequence contains 1000 rendered LDR perspective images with albedo, normals and depth groundtruth, and 1000 LDR panoramic images rendered in the corresponding viewpoint using path tracing. 
We use LDR perspective images as input and supervise albedo and geometry with paired GT. 
To supervise lighting, 
we sample adjacent panoramic images rendered at locations that are visible in the perspective input image. This makes the  environment maps lie in our region of interest, \ie in front of the camera. 
We follow the data split and preprocessing from \cite{srinivasan2020lighthouse}.
We use 90\% (1472) of the scenes to train our model and reserve 10\% (162) for evaluation.

When evaluating on real world data, we also finetune our model on IIW dataset~\cite{bell2014intrinsic} for albedo and NYUv2 dataset~\cite{Silberman:ECCV12} for depth and normal. 
We also collected 120 indoor LDR panoramas from the internet and jointly train lighting on these panoramas. 
More details are included in the Appendix.

\newcommand\hhh{1.05cm}
\begin{figure*}[t!]
\vspace{-2mm}
\centering
\setlength{\tabcolsep}{0pt}
\footnotesize
\begin{tabular}{cccccccccc}
\includegraphics[width=0.095\linewidth,height=\hhh]{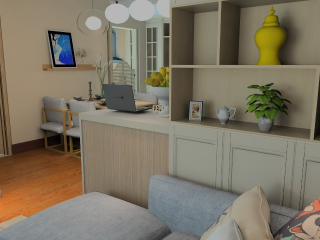} & \hspace{0.5mm}
\includegraphics[width=0.095\linewidth,height=\hhh]{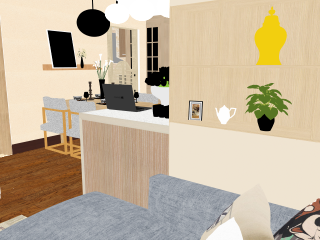} & 
\includegraphics[width=0.095\linewidth,height=\hhh]{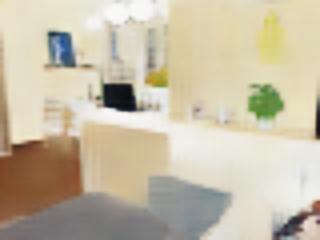} & 
\includegraphics[width=0.095\linewidth,height=\hhh]{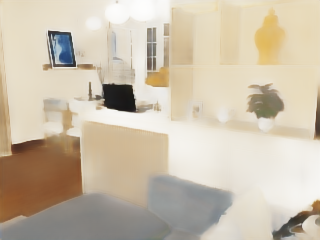} & \hspace{0.5mm}
\includegraphics[width=0.095\linewidth,height=\hhh]{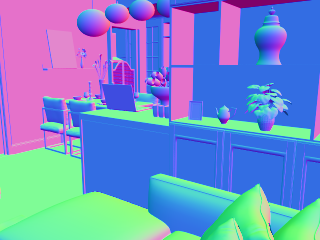} & 
\includegraphics[width=0.095\linewidth,height=\hhh]{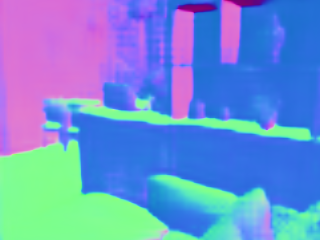} & 
\includegraphics[width=0.095\linewidth,height=\hhh]{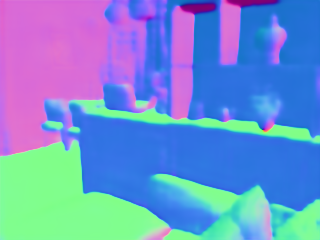} & \hspace{0.5mm}
\includegraphics[width=0.095\linewidth,height=\hhh]{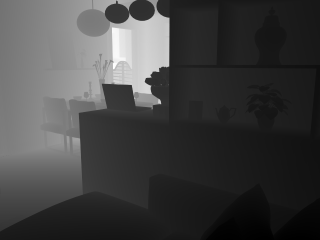} &
\includegraphics[width=0.095\linewidth,height=\hhh]{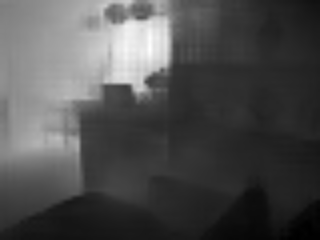} &
\includegraphics[width=0.095\linewidth,height=\hhh]{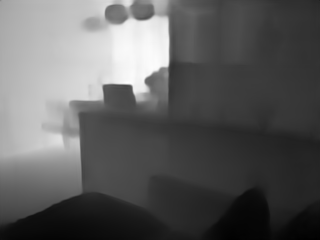} \\
\includegraphics[width=0.095\linewidth,height=\hhh]{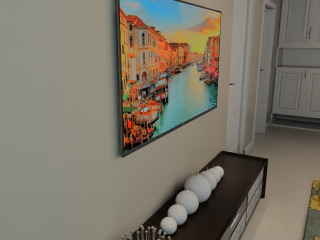} & \hspace{0.5mm}
\includegraphics[width=0.095\linewidth,height=\hhh]{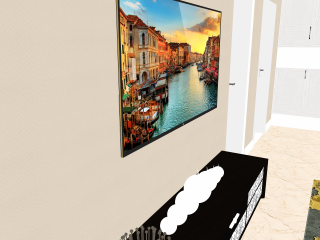} & 
\includegraphics[width=0.095\linewidth,height=\hhh]{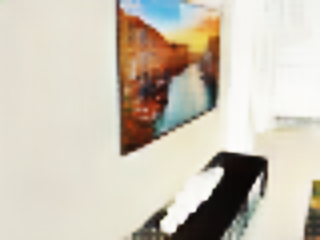} & 
\includegraphics[width=0.095\linewidth,height=\hhh]{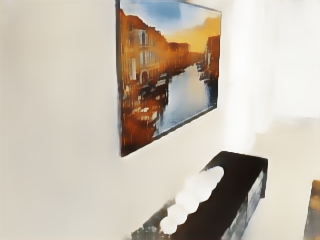} & \hspace{0.5mm}
\includegraphics[width=0.095\linewidth,height=\hhh]{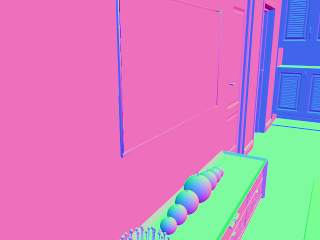} & 
\includegraphics[width=0.095\linewidth,height=\hhh]{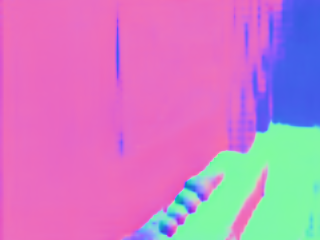} & 
\includegraphics[width=0.095\linewidth,height=\hhh]{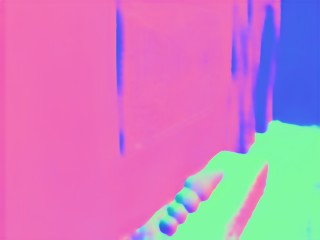} & \hspace{0.5mm}
\includegraphics[width=0.095\linewidth,height=\hhh]{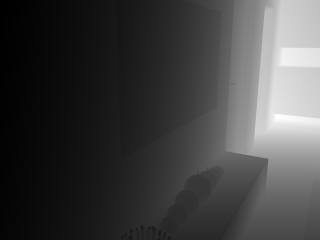} & 
\includegraphics[width=0.095\linewidth,height=\hhh]{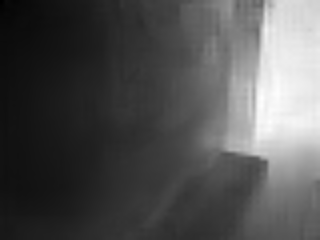} & 
\includegraphics[width=0.095\linewidth,height=\hhh]{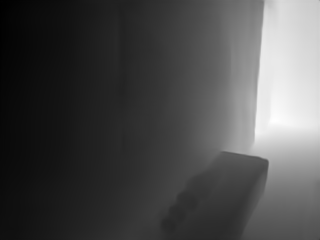} \\
Input image & GT Albedo  & Ours (w/o JR)  & Ours & GT Normal  & Ours (w/o JR)  & Ours & GT Depth  & Ours (w/o JR)  & Ours    \\
\end{tabular}
\vspace{-3.5mm}
\caption{\small \textbf{Qualitative results of predicted albedo, normals and depth.} The results are GT, our model without Joint Re-prediction (JR) Module and our full model. Joint Re-prediction enables joint reasoning and obtains crisper and more accurate results. }
\label{fig:qual_AND} 
\vspace{-4mm} 
\end{figure*}

\newcommand\hh{1.5cm}
\begin{figure*}[t!]
\centering
\setlength{\tabcolsep}{0.1pt}
\footnotesize
\begin{tabular}{cccccccc}
 & \multicolumn{2}{c}{Albedo} & \multicolumn{2}{c}{Normals} & \multicolumn{3}{c}{Re-rendered Image}    \\
\includegraphics[width=0.12\linewidth,height=\hh]{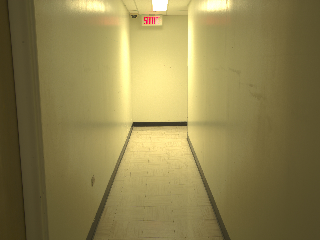} & \hspace{0.5mm}
\includegraphics[width=0.12\linewidth,height=\hh]{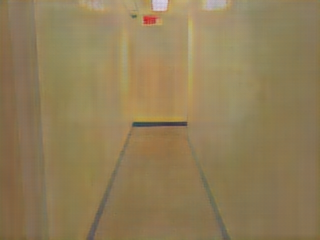} & 
\includegraphics[width=0.12\linewidth,height=\hh]{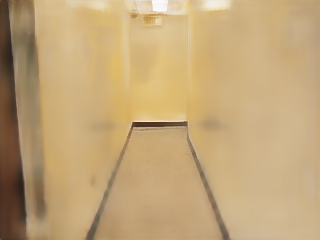} & \hspace{0.5mm}
\includegraphics[width=0.12\linewidth,height=\hh]{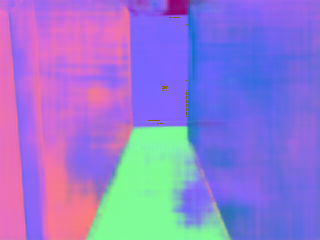} & 
\includegraphics[width=0.12\linewidth,height=\hh]{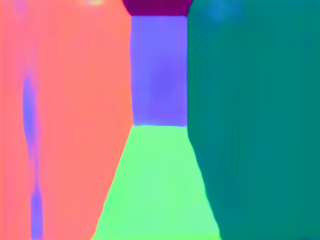} & \hspace{0.5mm}
\includegraphics[width=0.12\linewidth,height=\hh]{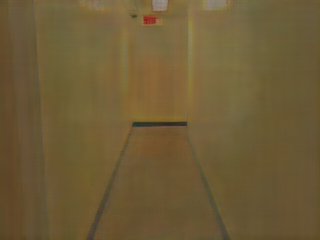} & 
\includegraphics[width=0.12\linewidth,height=\hh]{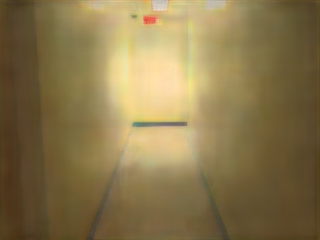} & 
\includegraphics[width=0.12\linewidth,height=\hh]{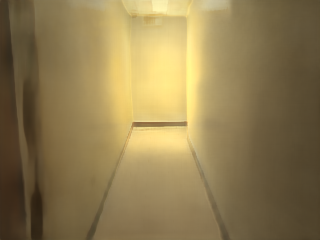} \\[-0.5mm] 
\includegraphics[width=0.12\linewidth,height=\hh]{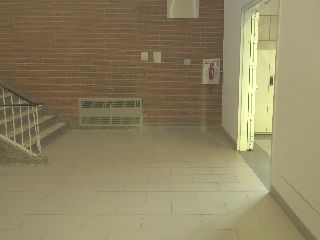} & \hspace{0.5mm}
\includegraphics[width=0.12\linewidth,height=\hh]{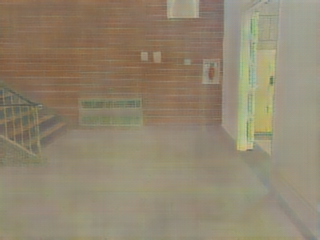} & 
\includegraphics[width=0.12\linewidth,height=\hh]{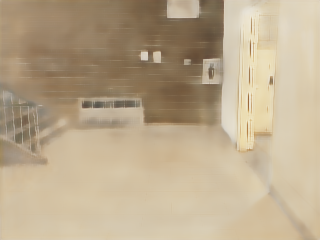} & \hspace{0.5mm}
\includegraphics[width=0.12\linewidth,height=\hh]{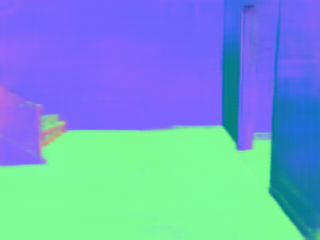} & 
\includegraphics[width=0.12\linewidth,height=\hh]{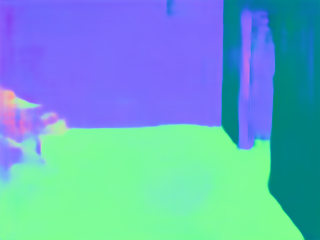} & \hspace{0.5mm}
\includegraphics[width=0.12\linewidth,height=\hh]{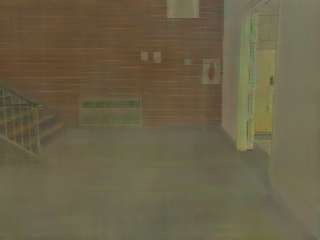} & 
\includegraphics[width=0.12\linewidth,height=\hh]{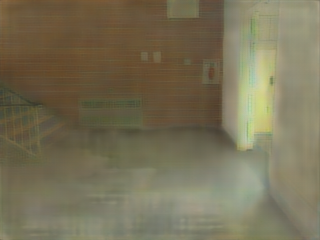} & 
\includegraphics[width=0.12\linewidth,height=\hh]{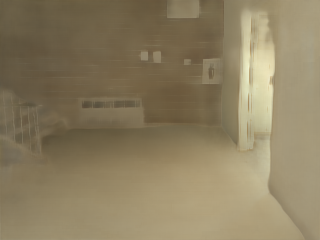} \\ 
Input image & NIR \cite{neuralSengupta19} & Ours  & NIR \cite{neuralSengupta19} & Ours & \tabincell{c}{NIR {\scriptsize (env.~map only)}} & NIR \cite{neuralSengupta19}  & Ours    \\
\end{tabular}
\vspace{-4mm}
\caption{\small \textbf{Qualitative comparison on predicted albedo, normals and re-rendered image.} 
Our fully physics-based lighting representation and differentiable renderer can better disambiguate and reproduce complex lighting effects with less artifacts. 
}
\label{fig:qual_rerender} 
\vspace{-5mm} 
\end{figure*}

\vspace{-4mm}
\paragraph{Evaluation. }
We evaluate albedo, normals and depth prediction on InteriorNet \cite{InteriorNet18} and real-world datasets IIW~\cite{bell2014intrinsic} and NYUv2~\cite{Silberman:ECCV12}.
For quantitative comparison, we use scale-invariant MSE (si-MSE) for albedo and depth due to scale ambiguity. 
We use mean angular error for normals and PSNR for lighting. 
We report the re-rendering MSE between re-rendered image and input image, which indicates whether the prediction is physically correct. 
The most effective lighting evaluation is through qualitative results.
We compare with prior works by visualizing object insertion results and the predicted environment map at a given location.

\vspace{-4mm}
\paragraph{Baselines. }
We quantitatively compare our methods with the state-of-the-art NIR \cite{neuralSengupta19} and the classic optimization-based method SIRFS \cite{barron2014shape}. In all experiment settings, we re-train NIR on the same data, \ie InteriorNet, to ensure a fair comparison. 
Li \etal \cite{li2020inverse} requires dense per-pixel lighting supervision and cannot train on the same data sources, and thus we provide qualitative comparison on the lighting prediction. 
For lighting estimation, we also compare with current state-of-the-art method Lighthouse \cite{srinivasan2020lighthouse}, which uses a stereo image pair as input instead of a monocular image.

\subsection{Evaluation of Albedo and Shape } 
\vspace{-1mm}
\paragraph{Evaluation on InteriorNet. }
We compare with baseline methods and ablate our model choices on InteriorNet. 
As shown in Table~\ref{table:interiornet_AND}, the performance of SIRFS exemplifies the limits of optimization-based methods on images of complex scenes. 
Our method outperforms NIR, indicating that our method better disambiguates the intrinsic properties. 
We also ablate our method. 
Predicted output from Direct Prediction Module is shown as ``Ours (w/o JR)''. 
Results show that the Joint Re-prediction Module helps improve the performance with the benefits of joint reasoning over the initial prediction. 
By comparing ``Ours'' and ``Ours (JR w/o lighting)'', the quantitative evaluation shows that the properties related to lighting 
(Eq.~\ref{eq:JR_lighting}) helps improve the overall performance. 
We provide qualitative results of albedo, normals and depth in Fig.~\ref{fig:qual_AND}. Results show that the Joint Re-prediction Module further disambiguates the intrinsic properties based on initial prediction and produces higher quality prediction. 

\vspace{-4mm}
\paragraph{Evaluation on real-world data. }
We evaluate albedo prediction on the IIW dataset \cite{bell2014intrinsic}, which provides sparse pairwise human annotations. We use the official Weighted Human Disagreement Rate (WHDR) as the metric, which measures the error when albedo prediction disagrees with human perception, and the results are shown in Table~\ref{table:iiw}. 
We also evaluate normal and depth prediction on the NYUv2 dataset in Table~\ref{table:nyuv2}. 
Our method outperforms baselines and validates the effectiveness of our Joint Re-prediction Module. 
Similar to prior works \cite{li2020inverse,neuralSengupta19}, we focus on the holistic inverse rendering framework and do not compete with state-of-the-art depth and normal estimation methods with specially designed architecture and other data sources \cite{hickson2019floors,lee2019big}. 

\begin{figure*}[t!]
\vspace{-3mm}
\centering
\begingroup
\setlength{\tabcolsep}{1pt}
\resizebox{0.999\linewidth}{!}{
\begin{tabular}{p{0.2\linewidth}<{\centering}p{0.2\linewidth}<{\centering}p{0.2\linewidth}<{\centering}p{0.2\linewidth}<{\centering}p{0.2\linewidth}<{\centering}}
\multicolumn{5}{c}{\includegraphics[width=1\linewidth,height=2.5cm]{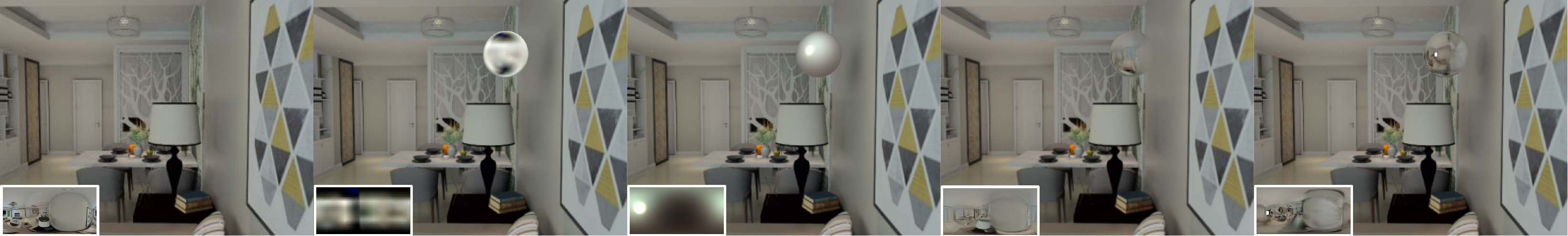}} \\[-0.5mm]
\multicolumn{5}{c}{\includegraphics[width=1\linewidth,height=2.5cm]{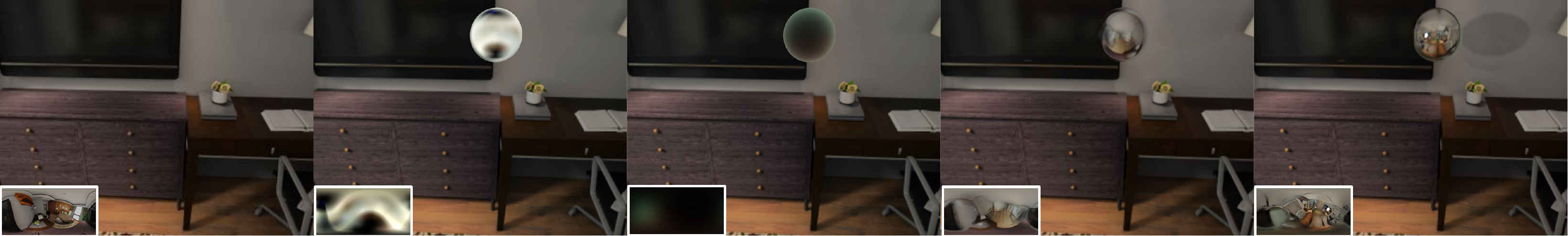}} \\[-1mm]
{\small Image \& LDR GT}  & {\small NIR \cite{neuralSengupta19}} & \small{Li \etal \cite{li2020inverse}} & \small{Lighthouse$^*$ \cite{srinivasan2020lighthouse}} & {\small Ours} \\
\end{tabular}
}
\endgroup
\vspace{-3.5mm}
\caption{\small\textbf{Qualitative comparison of lighting estimation. } We compare insertion of a \emph{purely specular} sphere, and on the bottom-left of each example displays the estimated environment map at the inserted location. Our method produces both angular details (env.~map) and realistic cast shadows with HDR, outperforming all competing methods. 
(* indicates use of a stereo pair as input. Best viewed zooming in. )
}
\label{fig:qual_L} 
\vspace{-2.5mm} 
\end{figure*}

\begin{figure*}[t!]
\footnotesize
\centering
\begingroup
\setlength{\tabcolsep}{0pt}
\resizebox{0.99\textwidth}{!}{
\begin{tabular}{ccccccc}
\includegraphics[width=0.13\linewidth]{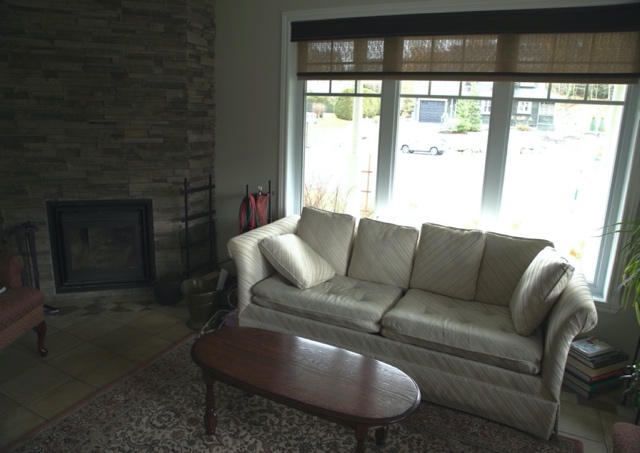} & \hspace{0.5mm}
\includegraphics[width=0.13\linewidth]{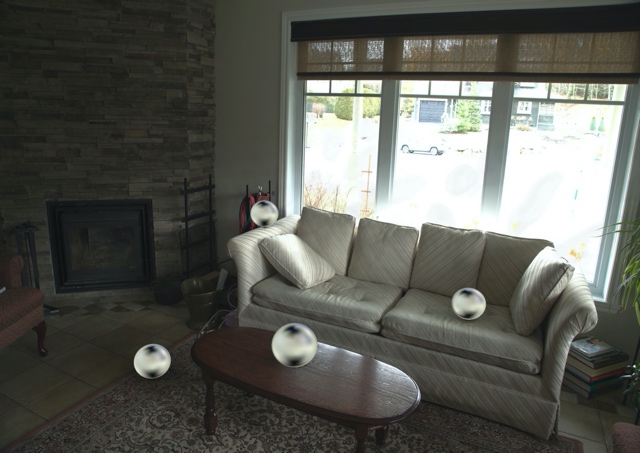} &
\includegraphics[width=0.13\linewidth]{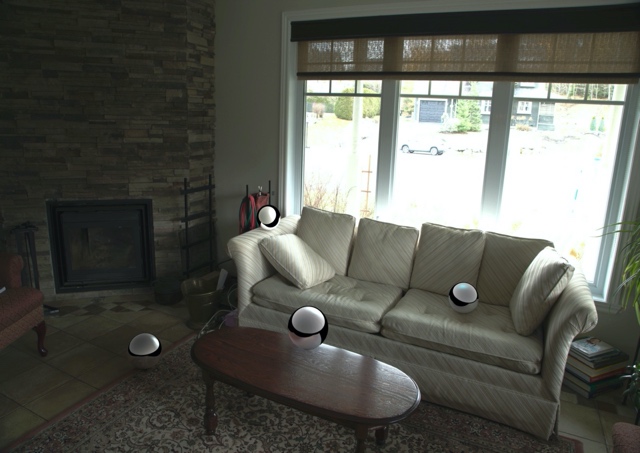} &
\includegraphics[width=0.13\linewidth]{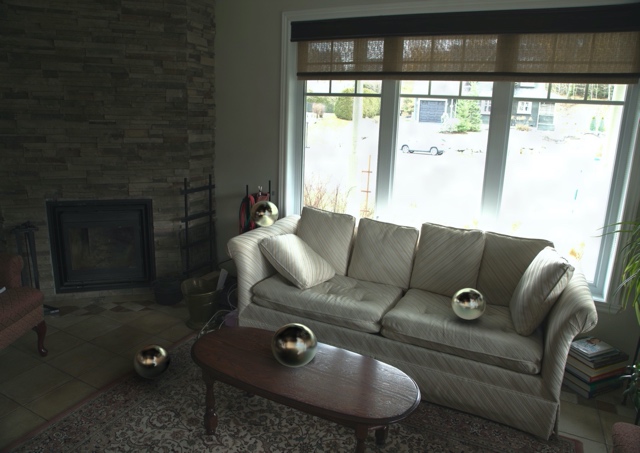} & \hspace{0.5mm}
\includegraphics[width=0.13\linewidth]{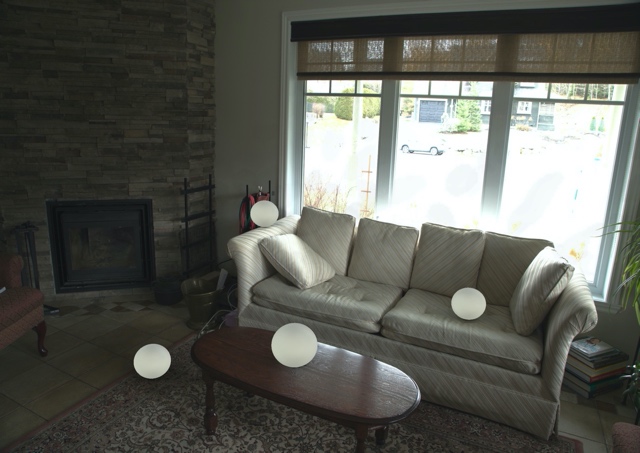} &
\includegraphics[width=0.13\linewidth]{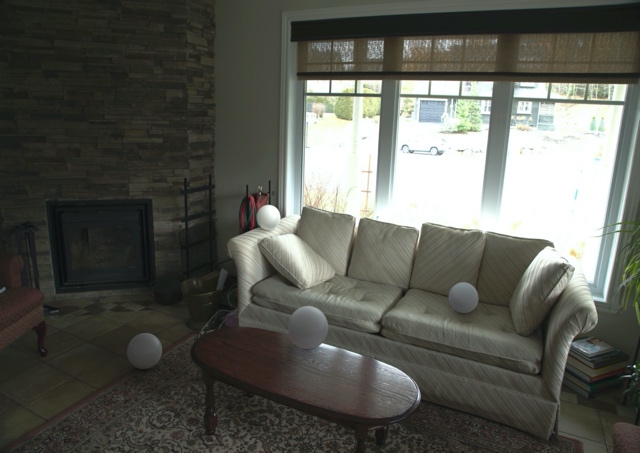} &
\includegraphics[width=0.13\linewidth]{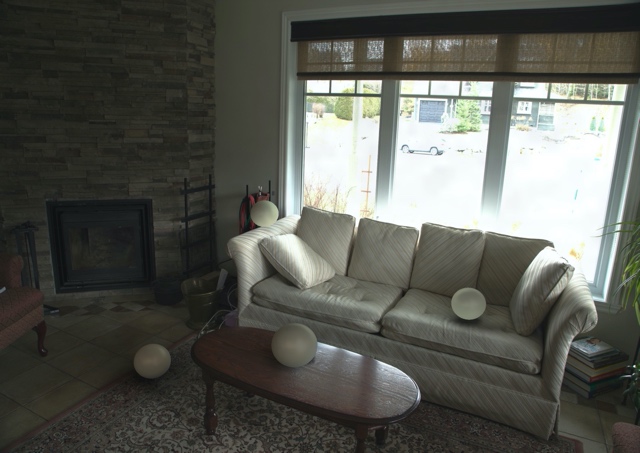} \\[-0.5mm]
\includegraphics[width=0.13\linewidth]{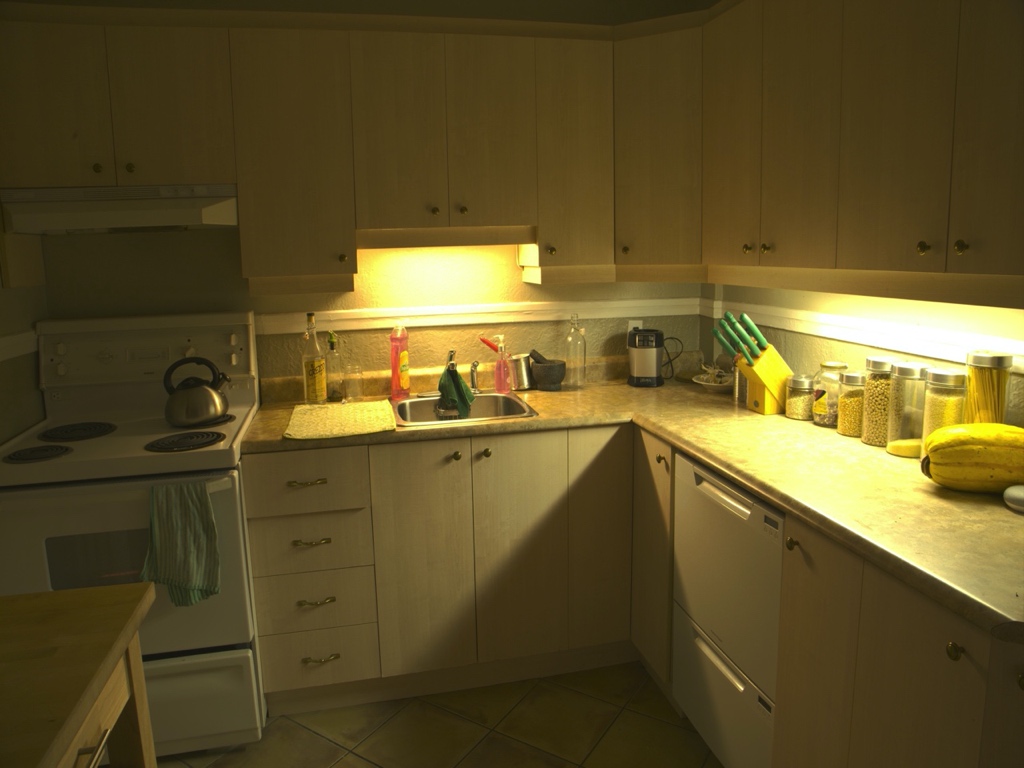} & \hspace{0.5mm}
\includegraphics[width=0.13\linewidth]{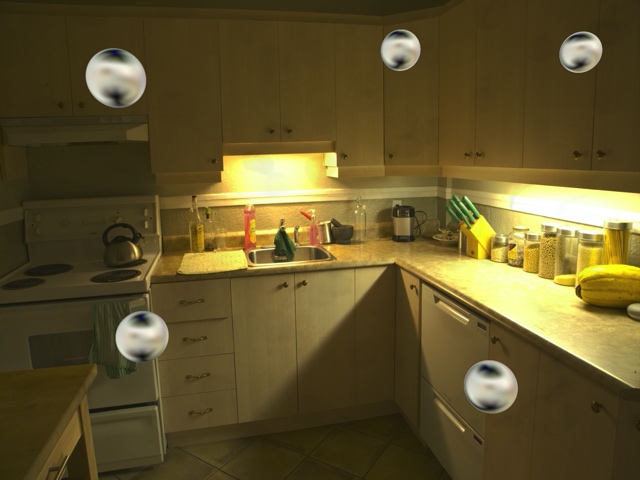} &
\includegraphics[width=0.13\linewidth]{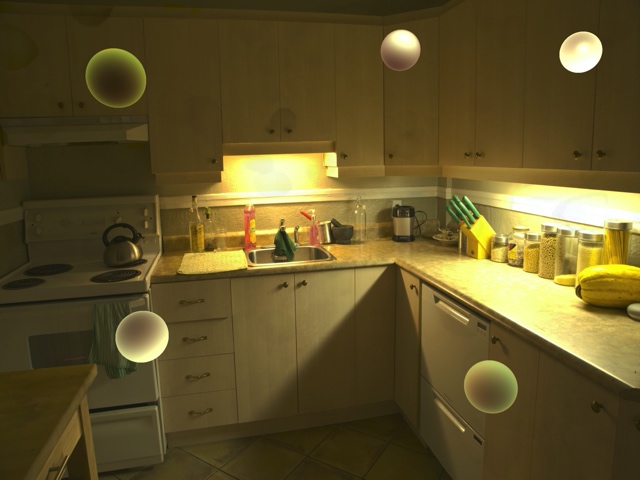} &
\includegraphics[width=0.13\linewidth]{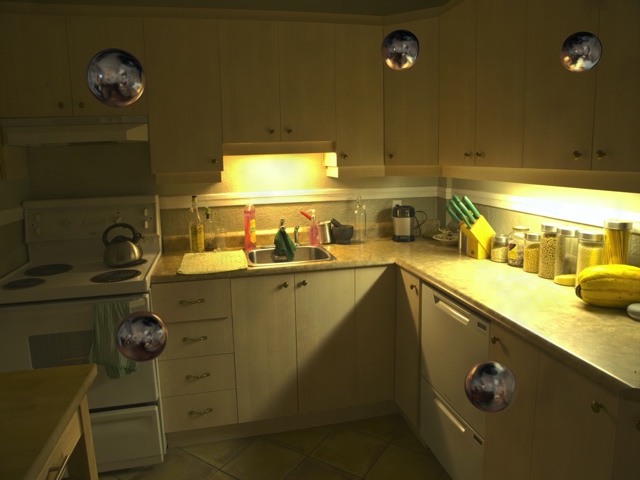} & \hspace{0.5mm}
\includegraphics[width=0.13\linewidth]{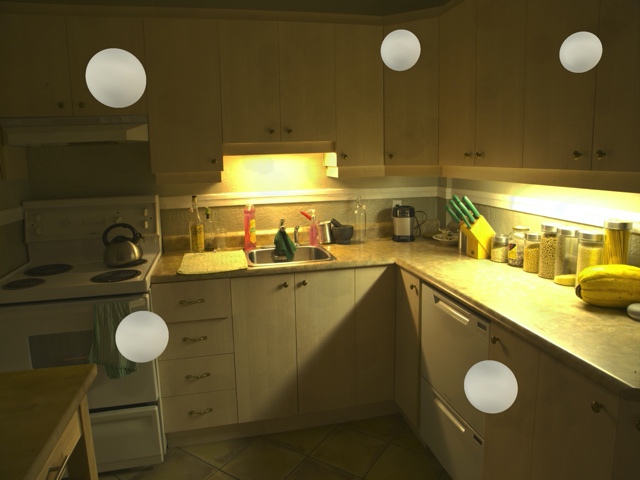} &
\includegraphics[width=0.13\linewidth]{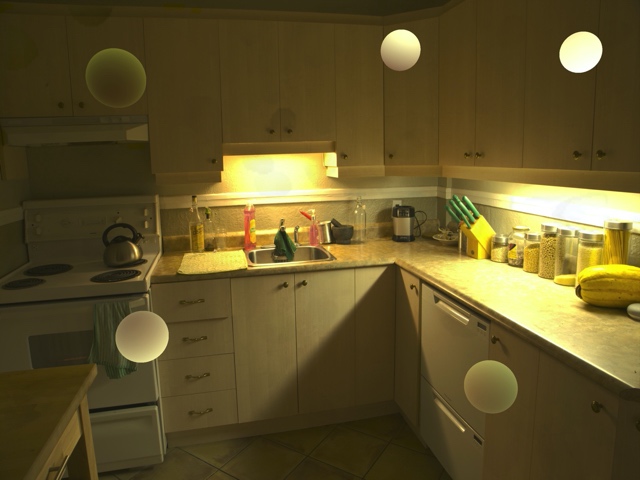} &
\includegraphics[width=0.13\linewidth]{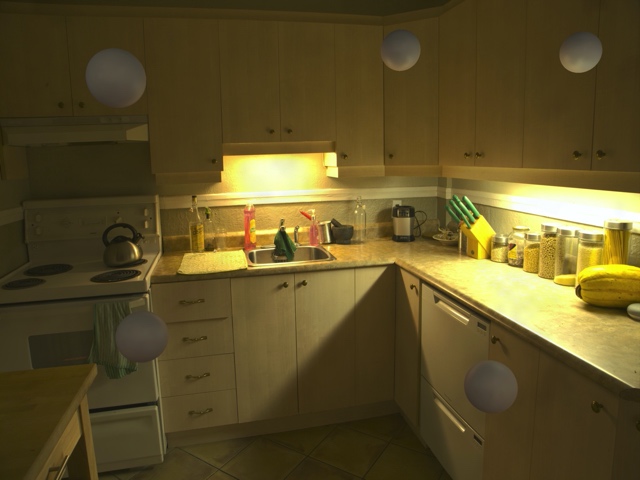} \\
Image  & NIR \cite{neuralSengupta19} & Li \etal \cite{li2020inverse} &  Ours & NIR \cite{neuralSengupta19} & Li \etal \cite{li2020inverse} & Ours \\
\end{tabular}
}
\endgroup
\vspace{-3.5mm}
\caption{\small \textbf{Qualitative comparison of lighting estimation on real-world images.} We compare \emph{purely specular} object insertion on the left, and on the right is mostly diffuse object. The top row shows insertion on a solid surface while bottom row shows freely inserted objects in 3D. Our method produces more realistic results in both specular and diffuse settings and is spatially consistent. 
(Best viewed zooming in. )
}
\label{fig:realworld} 
\vspace{-2.5mm} 
\end{figure*}

\begin{figure*}[t!]
\footnotesize
\centering
\begingroup
\setlength{\tabcolsep}{2pt}
\resizebox{0.998\textwidth}{!}{
\begin{tabular}{cccc}
\includegraphics[height=2.75cm,trim=0 0 0 140,clip]{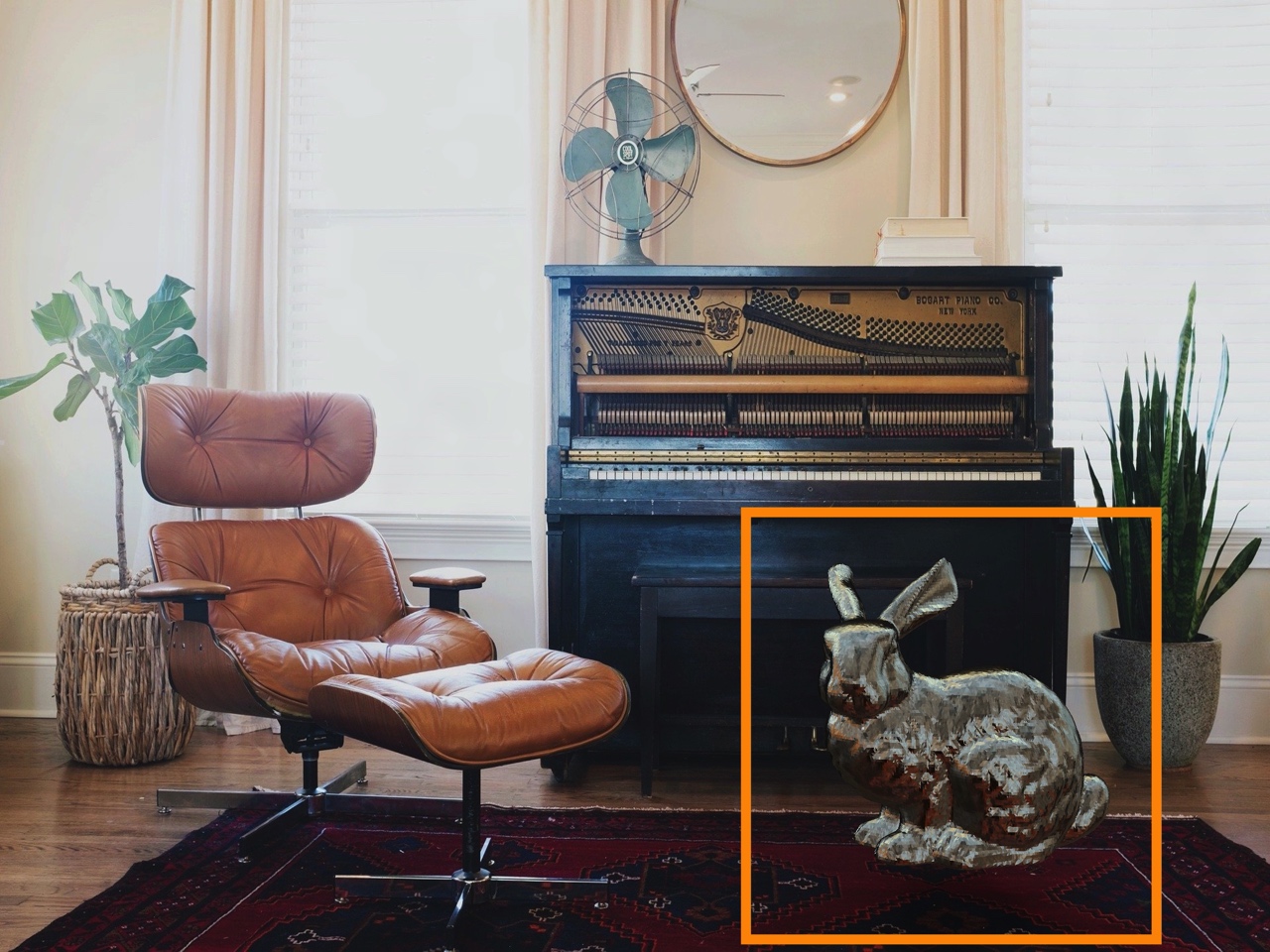} & 
\includegraphics[height=2.75cm,trim=0 0 0 90,clip]{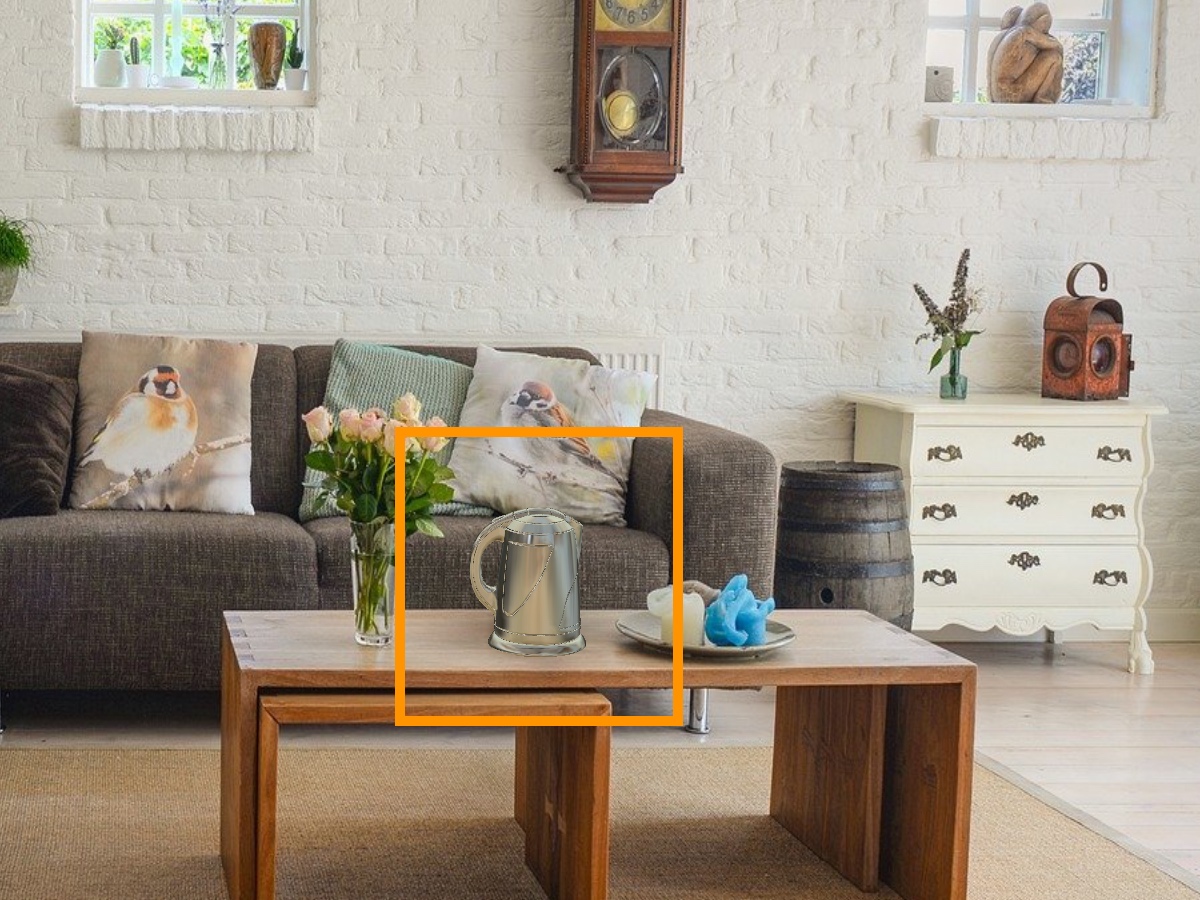} & 
\includegraphics[height=2.75cm,trim=0 0 0 160,clip]{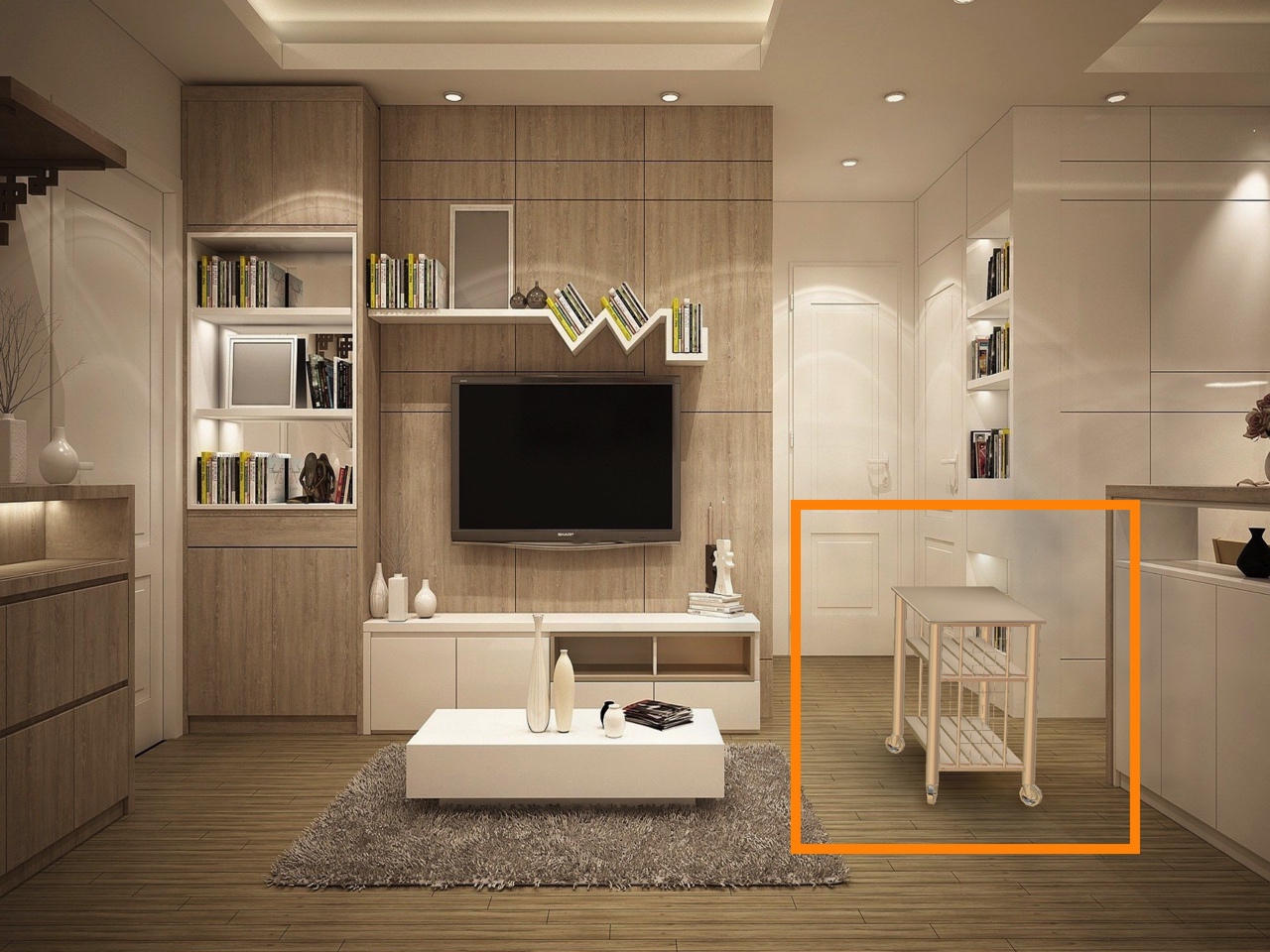} & 
\includegraphics[height=2.75cm,trim=0 0 0 180,clip]{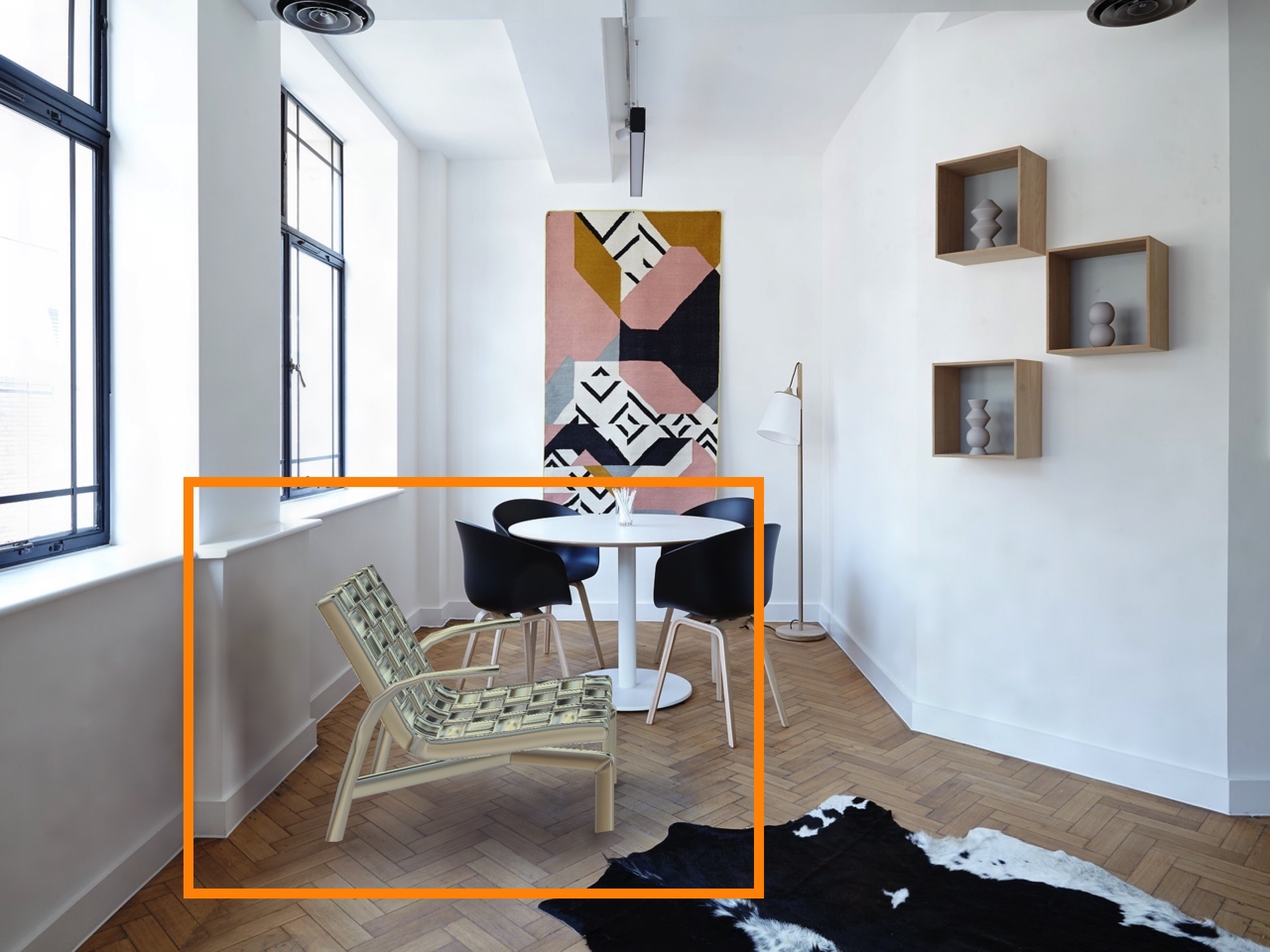} \\
\end{tabular}
}
\endgroup
\vspace{-4mm}
\caption{\small \textbf{Qualitative results of object insertion on real-world images.} From left to right, we insert a bunny, kettle, cart and armchair. }
\label{fig:object_demo} 
\vspace{-4.5mm} 
\end{figure*}

\vspace{-4.5mm}
\paragraph{Evaluation of re-rendered images. } 
We compare re-rendering error with baselines on InteriorNet and NYUv2 in Table~\ref{table:rerender}. NIR uses environment map as lighting representation, and employs a non-interpretable neural renderer, named residual appearance renderer (RAR), to account for all other lighting effects. 
Our method outperforms NIR both with and without the neural rendering module. 
We also show qualitative results of re-rendered images in Fig.~\ref{fig:qual_rerender}. 
Compared to the environment map re-rendering of NIR, we can handle 3D spatially-varying lighting and can re-render complex lighting effects, while NIR leaves these effects to the RAR module. 
Though addressing a more challenging task, our fully physics-based rendering process outperforms the RAR module which may easily produce artifacts and hurt the performance of other properties, such as artifacts in normals. 

We ablate our design choices in Table~\ref{table:rerender}. 
The re-rendering loss, which enforces joint reasoning of different properties, is critical for physically correct predictions. 
By comparing our Volumetric Spherical Gaussian with RGB$\alpha$ volume (Ours w/o SG), the results show that the spherical Gaussian volume increases the model capacity and leads to better re-rendering. 
We also tried jointly training on InteriorNet and 120 real-world LDR panoramas. We show that when evaluating on the real-world NYUv2 dataset, training on real-world LDR panoramas further improves performance.

\subsection{Evaluation of Lighting }
\vspace{-1mm}
\paragraph{Quantitative Evaluation. }
We evaluate our lighting prediction on InteriorNet in Table~\ref{tab:interiornet_L}. 
Our method significantly outperforms NIR \cite{neuralSengupta19} due to our 3D lighting representation that can handle spatially-varying lighting. 
Our method also outperforms Lighthouse \cite{srinivasan2020lighthouse}, the current state-of-the-art method for lighting estimation. 
Note that Lighthouse uses stereo pairs as input, which provides more information about depth and visible surface than monocular image. 
For the ablation study, empirical results indicate that our loss design is key in achieving the best performance. 
Compared to RGB$\alpha$ volume, our Volumetric Spherical Gaussian is able to capture view-dependent effects and brings better performance. 

\vspace{-4mm}
\paragraph{Qualitative Evaluation. }
We qualitatively compare the lighting estimation and virtual object insertion results with baselines in Fig.~\ref{fig:qual_L} on InteriorNet. Note that the inserted light probe is highly specular. 
NIR uses a single low-resolution environment map, which cannot handle spatially-varying effects and only recovers low frequency lighting, thus causing severe artifacts. 
Li \etal \cite{li2020inverse} employs 2D spatially-varying spherical Gaussian which can produce spatially-varying lighting, but the local lighting is still low frequency spherical lobes and cannot account for angular high-frequency details. These methods fail for inserting highly specular objects. 
Lighthouse \cite{srinivasan2020lighthouse} is using a volumetric RGB$\alpha$ lighting representation, which allows for 3D spatially-varying lighting. But the lighting volume in Lighthouse is learned with voxel inpainting from LDR panoramas, without supervision signal to make it \emph{physically correct}, and cannot predict HDR lighting. 
We can observe Lighthouse's lighting prediction has significantly less intensity variation, and the predicted LDR lighting cannot produce realistic cast shadows. 

Our method, with the Volumetric Spherical Gaussian lighting, produces more visually pleasing results with more realistic details. 
The re-rendering loss in our method enables joint reasoning and ensures that the model predicts physically correct HDR outputs due to the energy constraints. 
Our method is the only one that both preserves angular details and predicts HDR output for realistic cast shadows. 
Note the predicted cast shadows by our method are consistent with other visual cues in the scene (Top: lamp; Bottom: desk). 

We compare with prior works on real world data 
in Fig.~\ref{fig:realworld}. Here, we do not compare with Lighthouse as it requires stereo images. 
Comparing purely specular spheres on the left, our method preserves angular high-frequency details and is significantly more realistic. These benefits also apply to the diffuse spheres on the right. 
In the bottom row, spheres are from randomly sampled 3D locations. NIR \cite{neuralSengupta19short} uses a single environment map and can hardly capture spatial variation. Li \etal \cite{li2020inverse} uses 2D lighting representation, which cannot produce spatially consistent lighting and also cannot handle locations far from the 2D surfaces, leading to severe artifacts. 
Our method produces spatially coherent lighting with correct HDR intensity. 
We also show real-world object insertion results in Fig.~\ref{fig:object_demo}. 
Our method generalizes well to real-world images and consistently produces realistic shading and shadows. 
More results are included in the Appendix.

\vspace{-2mm}
\section{Conclusion}
\label{sec:conc}
\vspace{-2mm}

In this paper, we proposed a holistic monocular inverse rendering framework that jointly estimates albedo, normals, depth, and HDR light field. 
Our proposed Volumetric Spherical Gaussian representation nicely handles spatial and angular high-frequency details.  
With the physically based differentiable renderer, our method is capable of learning and reproducing the complex indoor lighting effects, and better disambiguate image intrinsics. 
Benefiting from joint training with the re-rendering constraint, our model can predict physically correct HDR lighting, despite being trained with only LDR images. 
We experimentally demonstrate that our model outperforms prior work on standard benchmarks, and is able to realistically render virtual objects in images with realistic shadows and high-frequency details, even when the objects are highly specular. 
We believe that these results demonstrate great promise of our model for AR applications.

{\small
\bibliographystyle{ieee_fullname}
\bibliography{egbib}
}

\end{document}